\documentclass[journal]{IEEEtran}
\ifCLASSINFOpdf
\else
\fi
\def\BibTeX{{\rm B\kern-.05em{\sc i\kern-.025em b}\kern-.08em
    T\kern-.1667em\lower.7ex\hbox{E}\kern-.125em}}

\usepackage{amsmath,amssymb,amstext} 
\usepackage[pdftex]{graphicx} 
\usepackage[dvipsnames]{xcolor}

\usepackage{tikz}
\usetikzlibrary{trees}
\usepackage{rotating, graphicx}
\usepackage{booktabs}
\newcommand{\tabitem}{\textbullet~~}
\usepackage{blindtext, xcolor}
\usepackage{comment}
\usepackage{pgfgantt}
\usepackage{lipsum}
\usepackage{dirtytalk}
\usepackage[numbers,sort&compress]{natbib}
\usepackage{caption}
\usepackage{subcaption}
\usepackage{svg}

\usepackage[compact]{titlesec}
\titlespacing{\subsection}{1pt}{*0.1}{*0.1}
\titlespacing{\subsubsection}{0pt}{*0}{*0}

\begin{document}
%
\title{{Autonomous Driving at Unsignalized Intersections: A Review of Decision-Making Challenges and Reinforcement Learning-Based Solutions}}



%

\author{Mohammad~Al-Sharman,~\IEEEmembership{Member,~IEEE,}
        Luc Edes,
        Bert Sun,
        Vishal Jayakumar,
        Mohamed A. Daoud,
        Derek~Rayside,~\IEEEmembership{Member,~IEEE},
        and William Melek,~\IEEEmembership{Senior~Member,~IEEE}

\thanks{M. Al-Sharman and D. Rayside are with the Department of Electrical and Computer Engineering and with Watonomous (Waterloo autonomous vehicle team in the SAE AutoDrive Challenge. (Watonomous.ca)), University of Waterloo, Waterloo, ON, N2L 3G1, Canada, e-mail: mkalsharman@uwaterloo.ca, drayside@uwaterloo.ca.}

\thanks{L. Edes is with David R. Cheriton School of Computer Science and with Watonomous), University of Waterloo, ON, CA.}%

\thanks{B. Sun is with the Faculty of Mathematics in the Department of Combinatorics and Optimization, and with Watonomous), University of Waterloo, ON, CA.}%

\thanks{Vishal Jayakumar, Mohamed A. Daoud, and W. Melek are with the Department
of Mechanical and Mechatronics Engineering and with Watonomous), University of Waterloo, ON, CA.}}%


%
%

\markboth{IEEE Journal Template ,~Vol.~--, No.~--, August~2023}%
{Shell \MakeLowercase{\textit{\emph{et al.}.}}: Bare Demo of IEEEtran.cls for IEEE Journals}
%



\maketitle

\begin{abstract}

Autonomous driving at unsignalized intersections is still considered a challenging application for machine learning due to the complications associated with handling complex multi-agent scenarios characterized by a high degree of uncertainty. Automating the decision-making process at these safety-critical environments involves comprehending multiple levels of abstractions associated with learning robust driving behaviors to enable the vehicle to navigate efficiently. In this survey, we aim at exploring the state-of-the-art techniques implemented for decision-making applications, with a focus on algorithms that combine Reinforcement Learning (RL) and deep learning for learning traversing policies at unsignalized intersections. The reviewed schemes vary in the proposed driving scenario, in the assumptions made for the used intersection model, in the tackled challenges, and in the learning algorithms that are used. We have presented comparisons for these techniques to highlight their limitations and strengths. Based on our in-depth investigation, it can be discerned that a robust decision-making scheme for navigating real-world unsignalized intersection has yet to be developed. Along with our analysis and discussion, we recommend potential research directions encouraging the interested players to tackle the highlighted challenges. By adhering to our recommendations, decision-making architectures that are both non-overcautious and safe, yet feasible, can be trained and validated in real-world unsignalized intersections environments.

\end{abstract}


%
\IEEEpeerreviewmaketitle

\section{Introduction}
%
%
%

%
\IEEEPARstart{F}{ollowing} the Defense Advanced Research Projects Agency (DARPA) Urban Driving Challenge (UDC), held in 2007, the research community has been encouraged to develop novel technologies to address technical and social challenges concomitantly with driving in urban settings autonomously \cite{buehler2009darpa,dikmen2016autonomous,wang2020ethical, al2020sensorless, kamal2021look,wang2020ethical1,zhou2023ralacs}. These challenges stem from the nature of urban driving itself, characterized by its complex multi-agent motion planning, in which the vehicle must react to various different scenarios including the interaction with other vehicles and traffic signals and signs \cite{kuwata2009real,verma2018vehicle,senanayake2020directional}. Unlike highway autonomous driving \cite{yang2023towards,tang2022highway,xu2022integrated}, driving in urban environments requires effective handling of complex multi-agent scenarios with a high level of uncertainty and occlusions \cite{li2015real,gilroy2019overcoming,huang2019uncertainty,dempster2022drg}. More specifically, driving at intersections is considered perilous for most human drivers. This can be justified by looking at the data reported in \cite{choi2010crash}. The Fatality Analysis Reporting System (FARS) and National Automotive Sampling System General Estimates System (NASS-GES) provide an estimation that 40\% of the crashes recorded in the US in 2008 occurred at intersections. They reasoned that among the factors contributing to these crashes, the most prevalent were related to the crash-involved drivers, namely, their age, gender and driving behaviour. Hence, deriving safe policies for autonomous vehicles that allow for safe crossing behaviour at intersections has been a topic of a profound importance as it can provide useful guidelines for designing preventive crash-mitigation schemes. Recently, academia and industrial partners have been extensively testing the most advanced autonomous technologies on their platform to ensure safe and efficient urban driving \cite{GM,Hu_2023_CVPR,chitta2022transfuser,chen2019autonomous,hawke2020urban}. 

Decision-making in autonomous vehicles is represented in a hierarchical complex structure \cite{schwarting2018planning}, covering a range of stages and sub-tasks which include the following: (i) planning where to go next, (ii) making decisions in the short and long term time frames based on the on-board sensors observations, (iii) decisions influenced by the interaction with other agents in the same environment, (iv) ensuring safe and robust vehicle control, (v) learning from the driving history and naturalistic human driving styles, (vi) coordinating with other vehicles to perform certain tasks collectively. However, within the context of urban intersections, enabling an autonomous vehicle to navigate safely and efficiently in such complex environments requires a high degree of autonomy, according to the Society of Automotive Engineers (SAE) J3016 standard \cite{TaxonomyAD}. Nonetheless, the current automated vehicles, even the fully autonomous ones, cannot fully navigate safely at all times, and cannot guarantee crash-free maneuvers due to critical decision-making errors \cite{khattak2020exploratory,wang2019crash}.

\begin{figure}
    \centering
    \begin{subfigure}{.23\textwidth}
        \centering
        \includegraphics[height=4cm, width=1\linewidth]{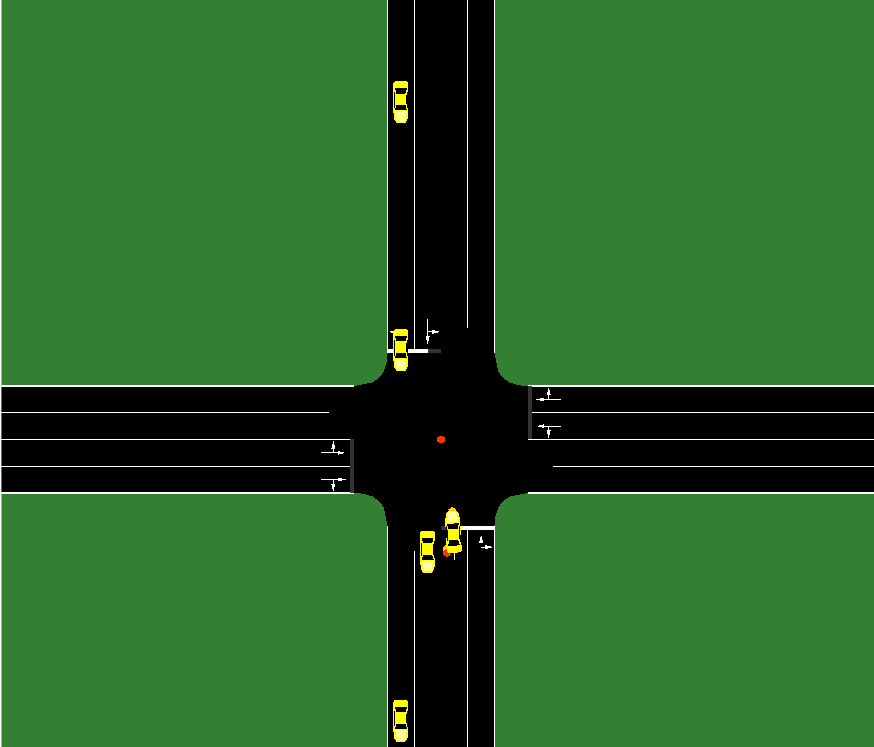}
        \caption{4-way}
        \label{fig:4way}
    \end{subfigure}%
    \hfill
    \begin{subfigure}{.23\textwidth}
        \centering
        \includegraphics[height=4cm, width=1\linewidth]{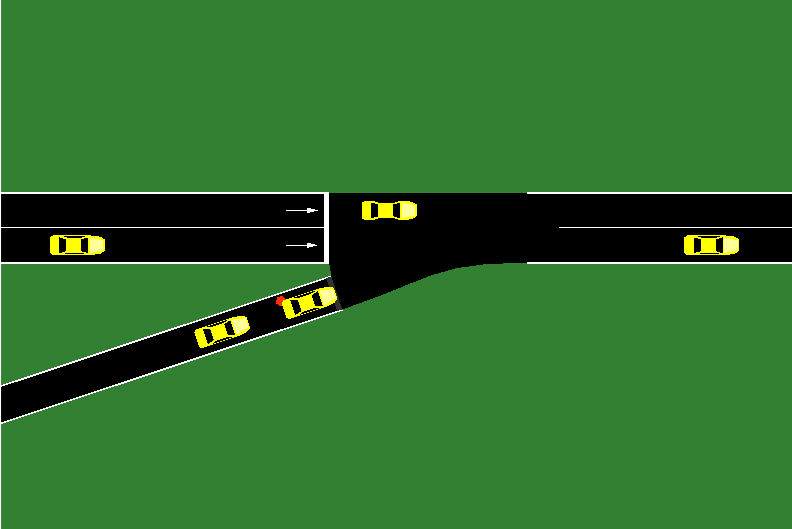}
        \caption{Junction merge}
        \label{fig:junctionmerge}
    \end{subfigure}
    \\
    \begin{subfigure}{.23\textwidth}
        \centering
        \includegraphics[height=4cm, width=1\linewidth]{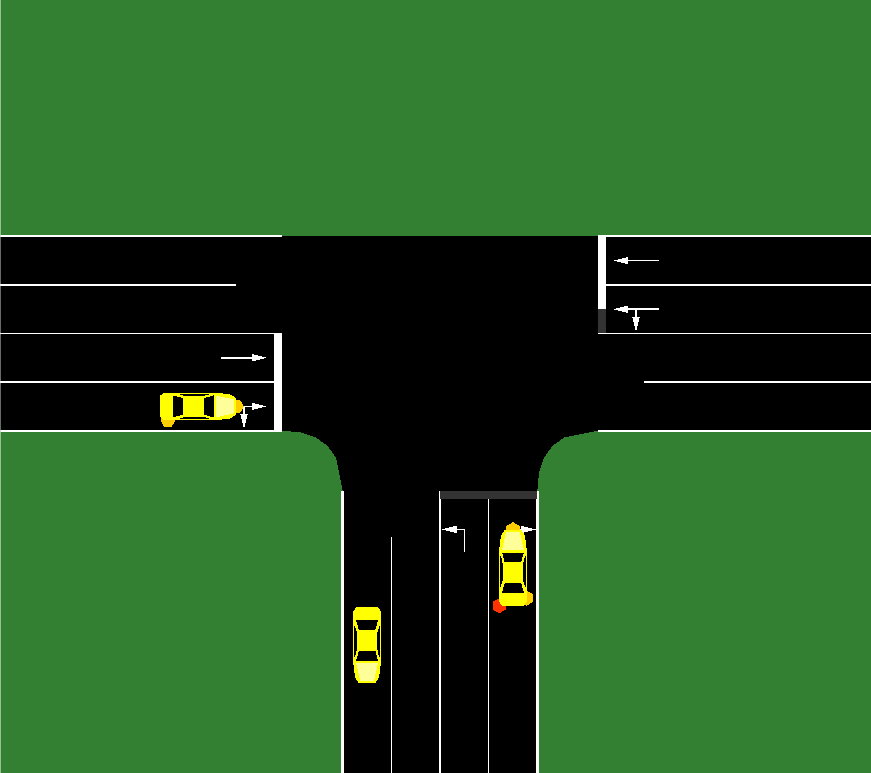}
        \caption{T-junction}
        \label{fig:tjunction}
    \end{subfigure}%
    \hfill
    \begin{subfigure}{.23\textwidth}
        \centering
        \includegraphics[height=4cm, width=1\linewidth]{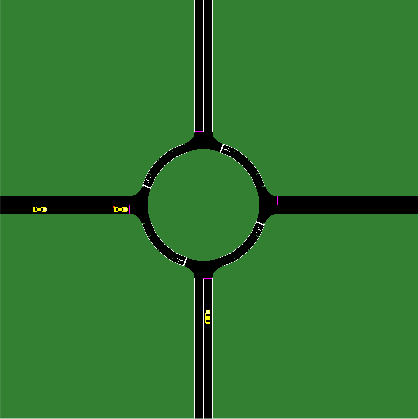}
        \caption{Roundabout}
        \label{fig:roundabout}
    \end{subfigure}
    \caption{Different types of unsignalized intersections. These images are generated using SUMO (Simulation of Urban MObility) traffic simulation software.}
    \label{fig:intersections}
\end{figure}

Making decisions at unsignalized intersections is a highly intractable process. The complex driving behaviour and the disappearance of traffic control signals make the motion inference of other intersection users highly-challenging (see Fig.\ref{fig:intersections}) \cite{pruekprasert2019game, li2020deep,huang2020diversitygan}. The non-stationary problem, along with the large partially-observable state space of agents dictate designing robust algorithms for safe intersection-traversal \cite{hubmann2018automated}. Numerous studies have been conducted to investigate algorithms to improve driving safety at unsignalized intersections. These algorithms have been introduced to tackle two main problems; inferring the intention of other intersection users and motion planning under uncertainty. We shall be discussing these algorithms with technical depth in the subsequent sections.

Based on our thorough investigation, we found that the proposed decision-making algorithms can be classified into three main categories: cooperative approaches, including game-theoretic, heuristic-based approaches and hybrid approaches which combine multiple classes of these algorithms for handling the unsignalized intersection problem. Cooperative approaches entail the use of V2V communication technology to exchange the states between the subject vehicle and other intersections users  \cite{hafner2013cooperative,wu2020cooperative,wang2021digital}. However, such technology is still an active area of research and has not been sufficiently developed to allow its application in existing decision-making schemes. Game-Theoretic-Based algorithms were adopted to model the vehicles' interactions in unsignalised intersections \cite{tian2020game, li2020game}. These game-theoretic based approaches assume that the states of the interacting vehicles are observed by the subject vehicle, which allows for predicting their future trajectories and then plan its own. However, this assumption is not likely to hold for current real-life decision making at unsignalized intersections. Heuristics-based approaches have been engineered to tackle safety-oriented problems associated with traversing urban intersections \cite{hubmann2017decision}. Researchers commonly classify these approaches into two main groups: rule-based and Machine Learning (ML) approaches \cite{qiao2018automatically}. Rule-based approaches use safety intersection metrics, namely time-to-collision (TTC), to generate distance-based traversing rules. However, engineering such rules to adapt with various possible crossing situations is a tedious process due to the large number of rules which need to be tuned. ML-based approaches, especially reinforcement learning approaches, focus on learning driving policies from the interaction between the vehicle and the intersection environment. 

\begin{figure}[t]
    \centering
    \includegraphics[height=11cm, width=7.0cm]{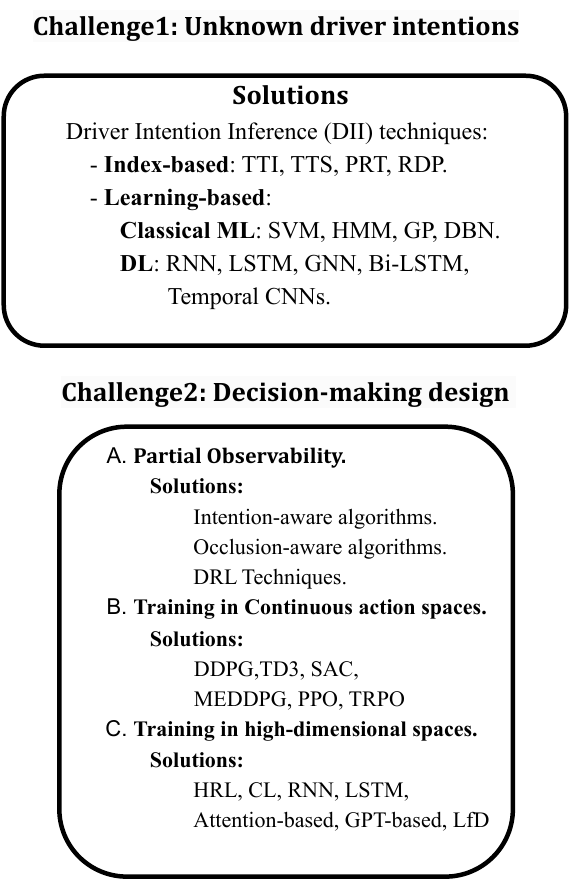}
    \caption{Surveyed decision-making challenges and solutions.}
    \label{fig:challenges}
\end{figure}

Applying modern RL-based approaches for learning optimal driving policies at unsignalized intersections has been studied extensively in the literature. Researchers have been motivated to develop these algorithms, owing to their capabilities in handling partially-observable environments by training its data-driven models based on mapping the environmental observations into actions \cite{silver2018general}. Nevertheless, design challenges behind developing crash-free intersection maneuvers and deploying them in real driving environments still need to be overcome. The surveyed schemes still suffer from several problems,i.e, the proposed design assumptions, the scalability of the proposed scheme to deal with more challenging urban driving scenarios, and the experimental validation in real urban driving settings. Hence, motivated by the published works, a review of the current and emerging trends in aspects related to decision-making in urban unsignalized intersections is recommended to lay the groundwork for potential advancement in this research direction. This survey offers an overview of algorithms and applications of decision-making in urban autonomous vehicles at unsignalized intersections, with the goal of continuing to explore methods to boost automation and enhance safety at these complex environments. Fig.\ref{fig:challenges} highlights the main decision-making challenges and their corresponding solutions surveyed in this paper.

\textcolor{black}{In comparison to existing survey papers on RL for autonomous vehicles, our review uniquely focuses on RL-based decision-making techniques specifically for unsignalized intersections-an area that has not been comprehensively covered in the literature. Previous works, such as those by Irshayyid et al. \cite{irshayyid2024review}, and Aradi \cite{aradi2020survey}, explore RL applications in highway control, and motion planning, respectively, but they do not address the unique challenges posed by unsignalized intersections. Other more recent surveys, like those by Chen et al. \cite{chen2024deep} and  Wu et al. \cite{wu2024recent} , focus on RL for path planning and for behavior planning, yet they also do not tackle the specific decision-making challenges inherent to unsignalized intersections. Additionally, a survey by Zhang et al. \cite{zhang2024survey} provides an overview of RL-based control for signalized intersections, which involves different complexities compared to unsignalized intersections. Our manuscript fills this critical gap by offering a detailed review of RL-based decision-making approaches for unsignalized intersections, highlighting their limitations, challenges, and potential future research directions. This contribution provides a fresh perspective that is valuable for advancing autonomous driving in complex, real-world scenarios.}

To define the theme of this survey clearly, we direct our attention towards various aspects related to behavioral motion planning for autonomous vehicles at unsignalized intersections. To be more specific, we focus this review on learning-based decision making schemes with a greater attention to algorithms that combine the recent advances of RL and deep learning for learning driving policies at unsignalized intersections. However, decision-making based on imitation learning or vehicle-to-vehicle (V2V) communication, in a connected driving fashion, is out of proposed survey's scope. Using V2V communications \cite{muller2022motion,li2021planning} can be a potential solution for anticipating vehicle behaviour and transferring it to the ego vehicle. However, for this solution to be fully viable, vehicular communication and connected vehicle technologies must be widely deployed. It should be noted that the vehicle-pedestrian interaction behavior is not covered in this proposed research. However, the reader is referred to a recent survey on  pedestrian trajectory prediction \cite{golchoubian2023pedestrian}.  

The main contributions of this survey paper can be stated as follows. First, an organised and in-depth state-of-the-art literature survey for decision-making at unsignalized intersections is proposed, highlighting the main navigational challenges and cutting-edge learning-based solutions. Second, an exploration of the Driver Intention Inference (DII) schemes at unsignalized intersections is carried out, with the goal of identifying key remarks for better handling the large partially-observable state space of the problem. Finally, based on the in-depth investigation, limitations of the published learning-based decision-making frameworks are identified and potential research directions are suggested to achieve better generalization characteristics of the trained traversing policies in real-life driving scenarios. 

To summarize, the rest of this paper is split into five sections as follows. Section II represents the background of this work, where we elaborate on aspects related to  decision-making in autonomous vehicles and reinforcement learning. Section III illustrates the challenges and learning-based solutions decision-making schemes with our observations on their inherent logic highlighted. Section IV presents remarks on possible future research directions. Finally, Section V concludes the proposed and future works.

\section{Background}

\subsection{Overview of decision-making in Autonomous Vehicles}

Autonomous vehicles (AVs) are considered autonomous decision-making systems as they provide continuous decisions based on processing perceptual observations. Along with these observations and sensor models, the predefined road network data, driving rules and regulations, dynamic behaviour of the vehicle, are utilized for predicting the vehicle's motion and generating low-level control commands autonomously. Developing such decision-making systems with a high degree of autonomy, is commonly organized by a well-defined multi-staged process \cite{schwarting2018planning}. However, the greatest challenges of motion planning in urban autonomous driving environments comprise the following factors: i) restricted sensing capabilities, specifically, vision and proximity in the  time-varying environment; ii) cluttering and occlusions in the scene which impede achieving accurate perception; iii) legal and technical constraints on the vehicle's response, arising from the driving rules and regulations.

\begin{figure}[t]
\centering
\captionsetup{justification=centering, skip=10pt}
\resizebox{1\linewidth}{!}{
\begin{tikzpicture}[node distance=4mm, >=latex',
       block/.style = {draw, rectangle, minimum height=10mm, minimum width= 6mm,align=center} ]

           \node [block]                         (Route)        {Route\\Planning};
           \node [block, right=of Route]         (behave)        {Behavioural\\Path Planning};
           \node [block, right=of behave]         (motion)  {Motion\\Planning};
           \node [block, right=of motion]   (local)   {Local\\Feedback control};

          \coordinate [above=of Route]         (input);

           \path[draw,->]  (Route)     edge (behave)
                           (behave)  edge (motion)
                           (motion)  edge (local);

\end{tikzpicture}}
\caption{Decision-making processes in urban autonomous vehicles.   \vspace{-2em}}
\label{Fig1}
\end{figure}
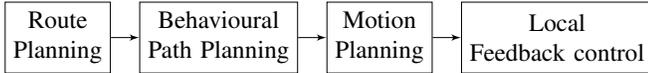

The hierarchy of the decision-making processes of urban autonomous vehicles is introduced in \cite{paden2016survey}. As shown in Fig.\ref{Fig1}, it consists of four cascaded layers, starting with the high-level route planning, followed by the behavioural and motion planning layers, and the low-level feedback control completes the scheme. At the very top layer, given the predefined destination, the autonomous decision-making system runs inherent route planning algorithms to compute the optimal path using the road network as a network graph. In this layer, the edge weights are summed in order to effectively solve for the routes with minimum cost. However, as the road network becomes larger, its graph network also becomes more complex making the use of the classical route planning schemes, namely Dijkstra \cite{broumi2016applying} and A* \cite{lavalle2006planning}, inefficient as the search time may exceed seconds \cite{liu2019intelligent}. Intelligent route planning approaches have been introduced for transportation efficiency enhancement. Advanced Deep Learning and Internet of Things (IoT) technologies have been employed for efficient route planning in complex urban transportation \cite{li2019traffic,ahmad2019real}. Once the optimal route has been defined, the next layer is focused on behavioral path planning. This layer is responsible for choosing proper driving behaviour based on the observed behaviour of other road drivers, traffic signals and road surface conditions. This behavior enables the AV to interact with road participants while performing lane changing, lane following, and other more complicated tasks like intersection traversal. For instance, choosing a cautious  behavior for intersection-traversal maneuvers based on the road conditions is a responsibility of the behavioral path planner. The planner will govern the approach, stopping, creeping and the intersection traversing maneuvers. Moreover, it will take into account the uncertainty associated with the intentions of other road drivers and pedestrians which may generate multiple possible trajectories. The problem of intention inference and its integration with the behavioral path planning schemes is discussed in greater detail in \ref{section:3}. After the driving behavior has been determined by the behavioral path planner, this behaviour has to be mapped into a vehicle's trajectory which will be tracked by the local feedback controller. Different aspects must be taken into account while choosing the proper trajectory (e.g. It must be feasible  with safety guarantees given the vehicle's dynamic model constraints). Finding such trajectory is an inherent component that must be accounted for by the motion planning layer. Finally, to execute this trajectory, a feedback controller must be designed to provide the correct input to govern the planned motion and compensate for the tracking errors arising from the assumptions made on the utilized vehicle dynamic model. For more details related to low-level feedback controllers, the reader is referred to the comprehensive survey conducted by Paden \emph{et al.} \cite{paden2016survey}.

\subsection{Modeling the decision-making problem}

Several research works have envisaged the decision-making problem at intersections as a reinforcement learning problem, where the agent and the environment interact continuously to learn an optimal policy that governs the vehicles' motion. The agent takes an action and the environment responds to this action and present new scenarios to the agent. A Markov Decision Process (MDP) is used to describe the environment for the RL problem \cite{qiao2018automatically}, where we assume that the environment for this specific problem is fully observable. Technically, MDP is described as a tuple $\{S, A, R, T, \gamma\}$ in which $S$ represents the observed states. These states may include information about the ego vehicle and states of other vehicles crossing the intersection. Among these states, velocities, position, and states related to the geometry of the intersection. $A$ and $T$ represents the set of actions and the transition function that maps state-action pairs to a new state. The immediate reward is defined by the reward function ${R}$, whereas ${\gamma}$ represents the discount factor for long-term rewards. 

In occluded intersections where the environment is not fully observed due to limited sensor range, occlusions in the scene, or uncertainty related to the pedestrians/ drivers intentions, a Partially-Observable Markov Decision Process (POMDP) is adopted to model these types of intersections. These cases shall be discussed in more detail in Section III.    

\subsection{Safety Assessment at Intersections}

At high-level decision-making, drivers perform safety assessment to avoid crashes and potential hazards. Shirazi \emph{et al.} \cite{shirazi2016looking} introduces five topics pertaining the safety assessment at intersections: Gap, Threat, Risk, Conflict and Accident. Gap assessment is an estimate used to anticipate the free distance between the leading and trailing vehicles. Gap distance-based and time-to-collision (TTC)  algorithms have been proposed for traversing intersection \cite{dresner2007sharing,alonso2011autonomous}. However, these simple approaches require laborious parameter-tuning to deal with different intersections scenarios. Given the locations of the subject vehicle, a threat assessment process is usually conducted to anticipate the potential threats of other road participants \cite{aoude2010threat}. In \cite{okamoto2017driver}, by inferring the intention of the road participants, threat predictions were obtained using random decision trees and particle filtering. A survey on the threat assessment technologies and state-of-the-art approaches can be found in \cite{li2020threat}. A risk assessment approach is used to detect risky scenarios which are related to the limited capabilities of the perception sensors or occluded environment which may result in incorrect decisions \cite{orzechowski2018tackling}. Risk assessment is usually coupled with predictions about the intention of other road participants. Intention-aware risk assessment has been done extensively to evaluate maneuvers at occluded intersections with limited perception capabilities. For detailed risk assessment at unsignalized intersection, the reader is referred to Section \ref{Section:II-B} which describes the state-of-the-art approaches of risk assessment for decision-making at urban intersections. Based on the environmental observations collected, conflict assessment is concerned with predicting the potential conflict scenarios of two or more vehicles that are going to collide if their movements remained unchanged \cite{jeong2019target}. Lastly, accident assessment is based on conducting precise analysis using data mining and machine learning techniques to make predictions that help in preventing crashes \cite{iranitalab2017comparison}. 

\subsection{RL Approaches}
\label{section2d}

\subsubsection{Preliminaries}
Reinforcement learning is a group of algorithms that focus at learning optimal policies via performing iterative experiments and evaluations for the sake of self-teaching overtime to achieve a specific goal. RL can be distinguished from other learning techniques such as supervised learning because the labels are timely delayed. The aim of RL is to learn an optimal policy $\pi$ which in charge of mapping the system states to control inputs that can maximize the expected reward $J(\pi)$. In eq. (\ref{eq:1}), the reward $r_{t}$ indicates how successful the agent was at a given time step $t$. For instance, large $r_{t}$ values are given when the agent is close to the desired trajectory, while small $r_{t}$ values are given when large deviations occur \cite{osa2018algorithmic}. The discounted accumulated reward is given as 
\begin{equation} \label{eq:1}J(\pi)=\mathbb{E}\left[\sum_{t=0}^{\infty} \gamma^{t} r_{t} \mid \pi\right]\end{equation}

The discount factor $\gamma$, where $\gamma$ $\in[0, 1]$, is used to adjust whether the agent is far-sighted or short-sighted. The desired policy can be described as 
\begin{equation}\pi^{*}=\arg \max _{\pi} J(\pi)\end{equation}

The value of the state $x$, is evaluated by calculating the expected return starting from $x$ and, subsequently, governed by policy $\pi$
\begin{equation}V^{\pi}(\boldsymbol{x})=\mathbb{E}\left[\sum_{t=0}^{\infty} \gamma^{t} r_{t} \mid \boldsymbol{x}_{0} =\boldsymbol{x}, \pi\right]\end{equation}

\noindent where $V^{\pi}\left(x_{t}\right)$ is defined by \cite{sutton2018reinforcement} as the \emph{value function}. Similarly, the action value in state $x$ is evaluated by calculating the expected reward starting from the action $u$ in a state $x$ and, subsequently, following policy $\pi$

\begin{equation}Q^{\pi}(\boldsymbol{x}, \boldsymbol{u})=\mathbb{E}\left[\sum_{t=0}^{\infty} \gamma^{t} r_{t} \mid \boldsymbol{x}_{0}=\boldsymbol{x}, \boldsymbol{u}_{0}=\boldsymbol{u}, \pi\right]\end{equation}

\noindent where $Q^{\pi}\left(\boldsymbol{x}_{t}, \boldsymbol{u}_{t}\right)$ is defined as the \emph{action-value function}. 

\textcolor{black}{Modern machine learning algorithms, driven by advancements in deep learning, have integrated Reinforcement Learning (RL) principles in the form of Deep Reinforcement Learning (DRL) \cite{wang2019deterministic,9144488}. Model-free RL learners are employed to sample the MDP to infer information about the unknown model. Given DRL’s strengths in handling partially observable environments with large state spaces and providing continuous action outputs, several DRL architectures have been developed to learn optimal policies for unsignalized intersection navigation \cite{de2018integrating,chen12019deep}. In this review, we examine these approaches in detail and discuss their specific applications in subsequent sections.}

\section{Unsignalized Intersection-Traversal: Challenges and Solutions }
\label{section:3}

 To enhance the AVs' ability to navigate complex urban unsignalized intersections, major technical challenges need to be investigated. In this section, we survey these challenges that arise while designing an automated learning-based decision-making algorithm for traversing these safety-critical environments.

 \subsection{Autonomous Driving Under Uncertainty}
 \label{section3a}
 
The uncertainty associated with motion prediction of other intersections vehicles at unsignalized intersection is caused by the following factors \cite{hubmann2018automated}:

 \begin{itemize}
  \item \textbf{Unknown intention of intersection users}.
     The motion of other intersection participants is highly connected to the future trajectory of the ego vehicle \cite{liu2019integrated}. Hence, for safe intersection navigation, precise motion predictions of the intersection users must be obtained. The main difficulty with inferring intention arises from the intrinsic uncertainty in the unknown current states and hidden variables, namely, unknown final destinations as well as their unforeseeable future longitudinal path \cite{wang2019trajectory}, and their likelihood of interaction with the subject vehicle \cite{trentin2021interaction}.   
  \item \textbf{Noise characteristics of sensors' observations}.
     The noise associated with the measurements collected from the mounted sensors adds another layer of uncertainty to the decision-making problem.
  \item \textbf{Occluded environments and limited perception}.
     The ability to observe the scene accurately is hindered by environmental obstructions and occlusions. \cite{yang2018scene}.  
\end{itemize}

\begin{figure}[!ht]
    \centering
    \includegraphics[height=7.7cm, width=8.4cm]{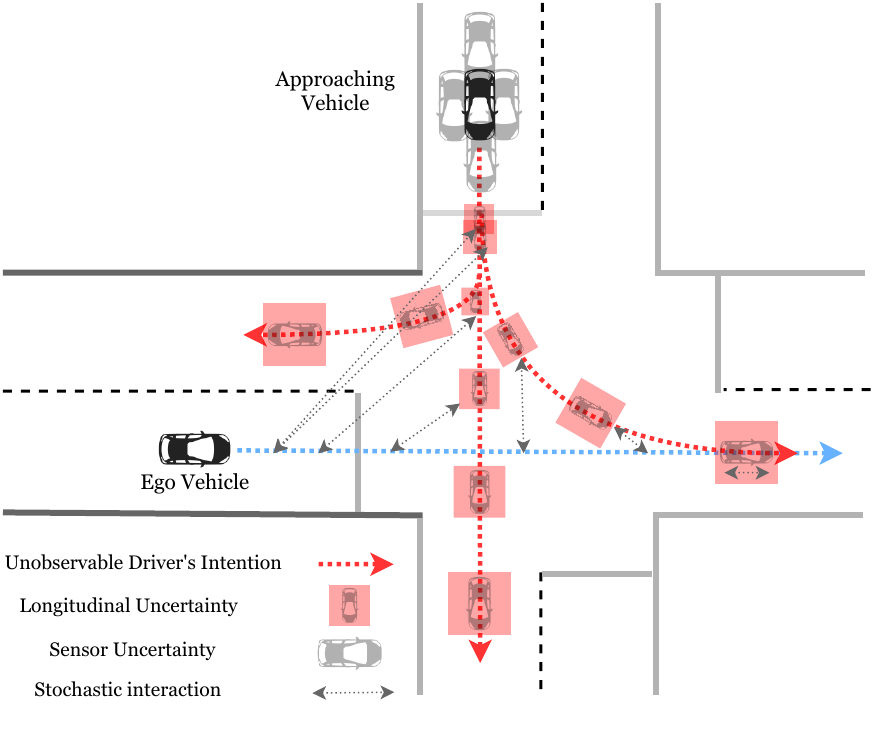}
    \caption{ An intersection-traversal scenario where the ego vehicle is required to handle several sorts of uncertainties associated with the approaching vehicle.}
    \label{fig:1.7}
\end{figure}

Fig.\ref{fig:1.7} depicts an illustrative example of where these uncertainties originate from at a four-way unsignalized intersection.
 
Considering these uncertainties when designing learning-based decision-making schemes in a complex intersection environment is essential for the ego vehicle to traverse intersections safely. For example, predicted motion and future trajectories of the target vehicles \cite{yoo2021virtual}, which share potential conflict points with the ego vehicle need to be incorporated while solving for an optimal traversal policy of the ego vehicle. This policy needs to be optimized for the most probable future scenarios coming from stochastic and interactive motion models of the other target vehicles. Considering these scenarios, in real time settings, these policies allow the autonomous vehicle to incorporate the estimated change in future prediction accuracy in the optimal policy \cite{hoel2020reinforcement}. This yields a compact representation with reduced-dimensions state-space

Based on our observation, we found that researchers have been mainly focusing at developing learning-based frameworks to tackle two main technical problems; inferring the intention of the intersection users and designing the decision process. Hence, in sections \ref{Section:II-B} and \ref{Section:III-c}, we focus on exploring the published works on these problems with greater details.  

\subsection{Driver Intention Inference Challenge}
\label{Section:II-B}
Accurately inferring and forecasting the intentions of drivers at unsignalized intersections is crucial for addressing the cause of an accident and ensuring road safety in such diverse multi-agent environments. Several research efforts have been exerted in order to develop algorithms for DII applications. These algorithms tackle the intention inference problem as a classification problem where intentions are classified based on the driving behaviour \cite{liu2019driving,trende2021modelling}. These DII approaches can be classified into two groups: index-based and learning-based. In index-based approaches, safety metrics are utilized to examine driving behaviors at intersections in order to formulate risk assessment schemes. For example, time-to-intersection (TTI), time-to-stop (TTS), time-to-collision (TTC), Perception Reaction Time (PRT), Required Deceleration Parameter (RDP), along with brake application were taken into account for inferring the driver's intent at intersections \cite{scanlon2016predicting,doerzaph2007development}. These index-based approaches, however, are designed for only frontal-crash prevention systems, where, in real driving scenarios, careless drivers may collide with the ego vehicle from different angles. Classical machine learning (ML) classification techniques have been also employed for intention inference applications. For instance, Aoude \emph{et al.} \cite{aoude2010threat} proposed a Support Vector Machine-based (SVM) intention predictor that was developed as part of the proposed threat assessor scheme. Subsequently, the developed threat assessor warns the host vehicle with the identified threat level and advises the best escape path. Hidden Markov Models (HMM) were implemented for intention inference along with Gaussian Processes which were used for collision risks prediction of multiple dynamic agents \cite{laugier2011probabilistic}. Lefevre \emph{et al.} \cite{lefevre2012evaluating} reported using a Dynamic Bayesian Network (DBN) for developing a probabilistic motion model where intentions are estimated from the joint motion of the vehicles.  However, these ML-approaches fall short as they cannot capture the long-term temporal dependencies in the data.

Motivated by their efficacy in modelling sequential tasks, researchers have employed deep-structured Recurrent Neural Nets (RNN) for determining the intentions of drivers at non-signalized intersection. Zyner \emph{et al.} \cite{zyner2017long} introduced the use of long short-term memory (LSTM) for intention inference at unsignalized intersections. Observations on the dynamic states, namely, position, velocity and heading states, were captured by the on-board set of sensors and used to train the network. In \cite{phillips2017generalizable}, a group of 104 features were utilized from the NGSIM dataset to train the proposed LSTM-based intention classifier. These features encompass ego position and dynamics, surrounding vehicles and their past states, and rule features that highlight what legal actions can be taken in the current lane. The proposed method demonstrated high classification accuracy for intention prediction at intersections with different lanes or shapes. However, these methods rely heavily on the mounted on-board positioning/tracking system. This means that tracking data from GPS and Inertial Measurement Units (IMUs) are required in order for the system to operate effectively, restricting their usage to vehicles where streaming from these sensors is available. Zyner \emph{et al.} \cite{zyner2018recurrent} proposed a solution to this problem by using data from a Lidar-based tracking system which will be implemented in future intelligent vehicles. The proposed model was validated using a large naturalistic dataset which was collected from two days of driving at an unsignalized roundabout intersection. Recently, Jeong \emph{et al.} \cite{jeong2020surround} proposed an LSTM-based architecture for predicting the target vehicle’s intention based on their estimated future trajectory at unsignalized intersections.  This network was developed to study long-term dependencies between vehicles in complex multi-lane turn intersections, and was based on the previous sequential motion of the target vehicles measured by the sensors equipped with the AV. The predicted target motion is integrated with Model Predictive Control (MPC) which is responsible for planning the motion of the subject vehicle. Girma \emph{et al.} \cite{girma2020deep} introduced the use of Bidirectional LSTM with an attention mechanism for intention inference at unsignalized intersections based on sequence-to-sequence modeling principles (i.e. Surrounding vehicles trajectory analysis with recurrent neural networks).  Bidirectional LSTM is used due to its capability for exploring information from previous and future time steps. However, the proposed method is agnostic to the decision-making problem. Thus, integrating the proposed method with  decision-making scheme in real-time format is a research direction to be explored. More recently, Pourjafari \emph{et al.} \cite{pourjafari2023navigating} proposed a LSTM-based intended exit predictor that predicts the intended exit path of surrounding vehicles as they approach the intersection, given their dynamic states, namely position, velocity and heading. The intended exit predictor is integrated with GNN-based models to predict the sequence of vehicles traversing each collision point in the intersection and the approximate time window the subject vehicle requires to cross the collision point.

Table \ref{table:2} summarizes the surveyed deep-learning-based intention inference schemes highlighting their research objectives and significant remarks.

\begin{center}
  \begin{table*}[t]
    \caption{Summary of the covered Deep-learning-based intention inference schemes in this section}
    \centering
    \begin{tabular}{ c|p{4.85cm} | p{2cm}  | p{4.85cm} }
     \toprule
      \textbf{Reference} & \textbf{Objective} & \textbf{Method} & \textbf{Remarks} \\ \hline
       \cite{zyner2017long} & Intention inference based on the ego vehicle’s observations i.e. GPS, IMU and Odometry).& RNN &\tabitem 100 \% classification accuracy on the Naturalistic Intersection Driving Dataset \cite{bender2015predicting}.\\ \hline
        \cite{phillips2017generalizable}& Intention inference at multi-lane intersection based on observations of speed, lanes and six adjacent vehicles. & LSTM & \tabitem 85\% accuracy at intersections with different types and shapes (NGSIM) dataset.\\ \hline
        \cite{zyner2018recurrent}& Intention inference for ego vehicles without tracking data (GPS, steering wheel encoding). & LSTM. & \tabitem 	The results indicate that networks fed with more history up to 0.6 seconds perform better.\\
        & & & \tabitem The provided model gives 1.3 sec prediction window prior to any potential conflict.\\ \hline
        \cite{jeong2020surround} & Intention inference based on GPS, Lidars and different types of cameras (front and round views). & LSTM & \tabitem Based on the prediction results, longitudinal motion planning with safety guarantees is proposed using MPC .\\ \hline
        \cite{girma2020deep}& Intention Inference based on focusing on important time-series vehicular data. & Bidirectional LSTM with attention mechanism.  & \tabitem Sequence to sequence modeling is performed to map the input sequence of observation to a sequence of predicted driver’s intentions.\\
        & & & \tabitem Achieved high accuracy on the NDS dataset. \cite{gadepally2013framework}\\ \hline
        \cite{khairdoost2020real} & Intention inference for maneuver prediction at intersection based real driving sequences including vehicle dynamics, gaze data as well as head movements. &  LSTM & \tabitem A prediction window of 3.6s  has been achieved on RoadLab dataset \cite{beauchemin2010roadlab}.\\
        \hline
        \cite{azadani2022novel} & Intention inference for Path prediction using dilated convolution networks in conjunction with a mixture density network (MDN) considering the temporal aspects of driving data. & Temporal CNN & \tabitem  Outperforms ML-LSTM and LSTM-FL in terms of accuracy and computational complexity on ACFR dataset.\\
        \hline
        \cite{pourjafari2023navigating} & Intended exit path prediction based on location and speed of vehicles at unsignalized intersection & LSTM & \tabitem Intended exit predictor achieved 94.4\% accuracy on INTERACTION dataset \cite{zhan2019interaction}. \\
    \hline
    \hline
    \end{tabular}
    \label{table:2}
  \end{table*}
\end{center}

\subsection{Decision Making Challenge}
\label{Section:III-c}

Owing to the strengths of deep-structured neural networks in handling large partially-observable state-action space, major research directions have been followed aiming to develop learning-based schemes for tackling problems related to traversing unsignalized intersections autonomously. In this section, we present the main design challenges involved in developing learning-based algorithms for decision-making under uncertainty, as well as a review of relevant state-of-the-art solutions, emphasising key observations and shortcomings. 

\subsubsection{Partial Observability}

In real multi-agent autonomous driving settings, the agents have incomplete information about the environment with which they interact. Therefore, designing a robust decision-making framework in such environments is considered an intractable problem. In practice, such problems are typically modelled as (POMDPs), in which a driving policy is learned to provide safe actions while accounting for the stochasticity inherent in the process of inferring intention and motion planning \cite{sarkar2017trajectory}. Numerous works address the problem of modeling the decision-making process of the partially-observable driving environments at unsignalized intersections. Brechtel \emph{et al.} \cite{brechtel2014probabilistic} models the decision-making problem for navigating an occluded T-junction intersection as a POMDP. Uncertainties of the driver's behavior and the limitations of the perception of the environment were taken into consideration while solving the continuous POMDP. Sezer \emph{et al.} \cite{sezer2015towards} develops a mixed observable MDP (MOMDP) model, which is a variant of POMDP, for intention-aware motion planning at a T-junction intersection under the uncertainty of drivers intentions. Along with the unknown intentions of other drivers, their unknown future predictions in the longitudinal direction and their interaction with the ego vehicle are modeled in the proposed decision scheme in \cite{hubmann2018automated}. The problem is formulated as a POMDP where the solution of the POMDP is a policy determining the optimal acceleration of the ego vehicle. However, the scalability of the proposed scheme to deal with unknown intentions of oncoming vehicles from multi-directions has not been addressed.

Inspired by the strengths of Deep Reinforcement Learning (DRL) approaches in learning driving policy without the necessity to learn the MDP model itself, several works have recently adopted these methods to solve the designed MDPs. For instance, Isle \emph{at el.} \cite{isele2018safe} proposed a safe reinforcement learning algorithm for left turn intersection-traversal using action prediction techniques. An optimal policy is trained using deep Q-learning to minimize disruption to traffic which is measured by traffic braking and maximize the distance to other vehicles. To solve for an optimal policy in such a multi-agent environment, the problem was formulated as a Stochastic Game. Deep Q-learning Networks (DQN) have been used for solving intersection crossing problems modeled by POMDPs \cite{li2019urban}. A thresholded lexicographic Q-learning scheme was adapted to the deep learning framework. This algorithm mimics human driving in some challenging scenarios where safety is prioritized over traffic rules and ride comfort. A factored MDP model was utilized instead of full MDP to mimic the human driver behaviour and to improve the data efficiency. A SUMO traffic simulator was then used as the simulation environment to validate the proposed experiment. Given the limitation of the Deep Q-Network, Bouton \emph{et al.} \cite{bouton2019safe} introduced an integration of the POMDP planning, model-checking and reinforcement learning to derive safe policies which can guarantee that the vehicle can traverse urban intersections under multiple occlusions and perception faults. Empirically, an ablation study was conducted showing that the proposed approach exhibits superiority over conventional DQN methods. A Deep Distributional Q-learning algorithm was proposed to deal with uncertainties associated with the variety of human driving styles \cite{bernhard2019addressing}. The algorithm generates risk-sensitive actions based on offline distribution learning and online risk assessment. During the offline distribution learning, the distributions of the risk-neutral and state–action return are learnt from unknown behavior type of a participant sampled from a known environment. While the learned behaviour is being executed, the action risks (collisions) are quantified using distortion risk metrics where the optimal action can be then selected. Hoel \emph{et al.} \cite{hoel2020reinforcement} introduced a method to evaluate the uncertain actions (decisions) made by the agent in an unsignalized intersection environment. A Bayesian reinforcement learning method using an ensemble of neural nets with Randomized prior Functions (RPF) \cite{osband2018randomized}, has been introduced to estimate the distribution of \emph{Q-}values which are then utilized to estimate the action values. This proposed scheme shows robustness in identifying highly uncertain actions within and outside the training set which helps in choosing the safest actions for safe intersection traversing maneuvers. However, these proposed approaches fall short in terms of the proposed hard assumptions and the tailored intersection-traversal scenarios.

\begin{figure}[!ht]
    \centering
    \includegraphics[height=6cm, width=8.6 cm]{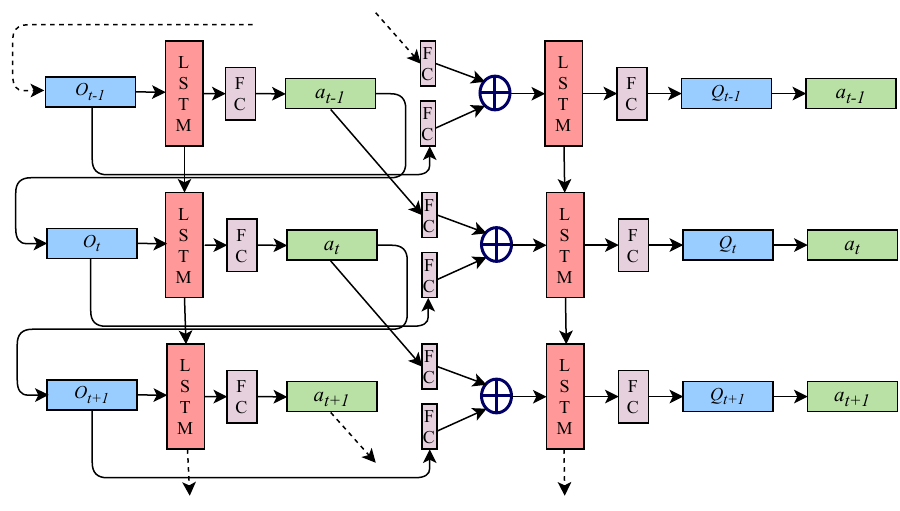}
    \caption{LSTM for solving the formulated POMDP of intersection-traversal problem.}
    \label{fig:3}
\end{figure}

The development of robust DRL algorithms for better handling of POMDP problems has piqued the interest of many researchers in the field. Zhu \emph{et al.} \cite{zhu2017improving} introduced a scheme called Action-specific Deep Recurrent Q-Network (ADRQN) to improve the learning capability in partially-observable environments. A fully connected (FC) layer is utilized to encode actions which are coupled with their corresponding observations to form action-observation pairs. LSTM is then adopted to process the time series of action-observation pairs. Similar to the conventional DQNs settings, the FC layer calculates the Q-Values based on the latent states learnt by the LSTM network. Another LSTM-based Deep Recurrent Policy Inference Q-networks (DRPIQN) was also introduced to handle partial observability caused by imperfect and noisy state information in real-world settings \cite{hong2017deep}. Both ADRQN and DRPIQN networks outperform other deep Q-learning techniques in terms of learning capabilities and stability when applied to a number of games. As an application to unsignalized intersection, Qiao \emph{et al.} \cite{qiao2018pomdp} proposed a network based on the design concepts of ADRQN and other deterministic gradient policy approaches for generating continuous time actions from the previous observations of the earlier steps. Observations from the previous 20 steps were used as inputs for the LSTM Network. Figure \ref{fig:3} exhibits the developed LSTM-Network which handles the POMDP and represents the decision-making problem of a four-way stop unsignalized intersection. The action output for each time step is obtained based on the observation inputs to the first LSTM and FC layers of the network at each individual time step. Subsequently, the Q-values are generated by taking the action of the previous step $a_{t-1}$ along with observation of the current step $O_{t}$ as an input to the second LSTM and FC layers. However, these approaches are entirely model-free as they rely heavily on the LSTM network to remember the past instead of having true belief states. Igl \emph{et al.} \cite{igl2018deep} proposed a Deep Variational Reinforcement Learning approach (DVRL), which relies less on a black box than the aforementioned DRPIQN and ADRQN, for learning optimal policies for POMDPs. Applying DVRL concepts for learning driving policy in partially-observable unsignalized intersection environments is still an area of research to be explored.

\textbf{Intention-aware schemes}. These probabilistic decision-making algorithms were developed to control the motion under the unknown intentions of intersection participants. For instance, a continuous Hidden Markov model (HMM) was developed to infer the high-level motion intentions including turning and continuing straight \cite{song2016intention}. A POMDP was then designed for the general decision-making framework, with assumptions and approximations used to solve the POMDP by calculating a policy to perform the optimal actions. Online solvers have also been used to solve the formulated POMDPs of the decision-making. In \cite{shu2021autonomous}, an improved variant of the POMCP solver which is called The Adaptive Belief Tree (ABT) is used to solve the proposed POMDP of an intention-aware left turning problem. The proposed decision-making problem is based on mimicking the human behavior of creeping slowly, upon reaching the stop line, to better understand the driver’s intention. The left-turn trajectory is simply assumed as a straight line with quarter circle curve. However, the intentions of the oncoming vehicles from only one direction is taken into consideration. In \cite{barbier2018probabilistic}, the uncertainties associated with human behavior of other drivers on the road in the context of an intersection have been modeled as a POMDP. An online solver has been utilized to find the optimal action that can be taken by the vehicle to react to uncertain situations \cite{barbier2018probabilistic}.  However, aside from a lack of real-world experimentation, using online solvers to solve is impractical because they only work for relatively small state spaces, and the complexity of solving the POMDP scales fairly quickly. Deep Reinforcement Learning, on the other hand, can work with much larger, or even large and continuous spaces, such as Atari \cite{igl2018deep}. 

Recently, Wang \emph{et al.} \cite{wang2023learning} employ Monte Carlo simulations within a Deep Reinforcement Learning (DRL) framework to replicate human decision-making at unsignalized intersections. The DRL agent, learning from diverse scenarios, is aided by a detailed environment model that includes state uncertainty, behavior modeling, and intention estimation for traffic participants like vehicles, pedestrians, and cyclists. This model accounts for varying driving styles and uses the Intelligent Driver Model \cite{Treiber_2000} for movement and intersection policies. By integrating human-like driving behaviors, the autonomous system adapts to various real-world conditions. The study also features a behavior cloning technique from human driving data, using Monte Carlo-generated training data to train a linear model for action quality prediction. This approach balances efficiency and safety, offering more interpretability than some neural network methods. The framework's effectiveness in achieving human-like behavior at unsignalized intersections is validated through simulations.

\textbf{Occlusion-aware schemes}. As previously mentioned in \ref{section3a}, due to environmental uncertainties and the limited capabilities of the sensors on the autonomous vehicle, occlusions can pose significant challenges to safely traverse an urban unsignalized intersection. Hence, many research papers have addressed this problem while integrating risk assessors into the decision-making schemes. For example, an occlusion-aware algorithm for left turn maneuver risk assessment at four-way unsignalized intersections was developed \cite{yu2019occlusion}. A particle filter paradigm was utilized to represent the distribution of the possible unobserved potential locations (particles) of the vehicle. However, this algorithm is not representative of how the vehicle can make decisions, but can be coupled with any POMDP or any other decision-making algorithm. The same group, based on the forward and background reachability, developed a probabilistic risk assessment and planning algorithm for a four-way intersection. The algorithm borders the risk-inducting regions arising from the occlusions of the ego-perception sensors that can be used to generate collision-free routes \cite{yu2019risk}. However, none of these algorithms were tested in real-world environments. McGill \emph{et al.} \cite{mcgill2019probabilistic} addressed the problem of navigating unsignalized intersections in the presence of occlusions and faulty perception \cite{mcgill2019probabilistic}. A risk model was proposed to assess the unsafe (risky) left turn across traffic at an intersection. Their model accounts for the traffic density, sensor noise and physical occlusions that hinder the view of other vehicles. By representing the intersection as a junction node with lanes entering and exiting the node, the risk assessment is used to determine a ‘go’ and ‘no-go’ decision at an intersection. The risk is modeled by defining near-miss braking incidents, collision incidents, traffic conflicts and small gap spacing. The risk is defined as the expected number of incidents that will occur if the ego-car enters an intersection. The overall risk is the sum of all expected incidents in all lanes and for all segments. In \cite{isele2018navigating}, the occluded intersection-traversal problem was viewed as a reinforcement learning problem. A deep Q-learning approach was utilized to traverse a partially observed four-way intersection. A creeping behavior upon reaching the intersection is learnt where the agent must perform an exploratory action to better comprehend the environment. Three action representations were studied: Time-to-Go, Sequential Actions and Creep-and-Go. In the Time-to-Go representation, the desired path is known for the agent, and the agent decides whether to go or to wait at every point in time. While in the latter scenarios, the agent can determine whether to accelerate or decelerate progressively. However, a bird’s eye view image space is used to describe the position of the vehicles in the space using Cartesian coordinates. This makes the implementation of the proposed DQN method inefficient for urban driving environments which require continuous actions rather than discrete ones. Moreover, in real AVs, it is infeasible to have a "bird’s eye view'" for acquiring the vehicle's state for decision-making applications. 

Table \ref{table:3} summarizes the major classes of decision-making schemes under partial-observability.  

\begin{center}
  \begin{table}[t]
    \caption{Classes of decision-making schemes under  partial-observability at unsignalized intersection}
    \centering
    \begin{tabular}{ c|p{3.5cm}|c}
     \toprule
      \textbf{ Class } & \textbf{Contribution} & \textbf{References}\\ \hline
      
        Occlusion-aware & \tabitem Navigation through static and dynamic occlusions.& \cite{lin2019decision} \cite{qiao2018automatically} \cite{brechtel2014probabilistic} \cite{kamran2020risk} \\  
        
           & \tabitem Navigating under perception errors due to occlusions & \cite{bouton2019safe} \cite{mcgill2019probabilistic} \cite{isele2018safe} \\  
          
           & \tabitem Navigating with Limited sensor range & \cite{naumann2019safe} \cite{ qiao2018pomdp} \cite{yu2019occlusion} \\ 
           
           & \tabitem A creeping behavior is learnt to better comprehend the environment.  & \cite{isele2018navigating} \\ \hline
        
        Intention-aware & \tabitem SVM-based motion planning.& \cite{jeong2019svm}\\

           & \tabitem Target motion-based behavioral planning. & \cite{jeong2019target}\\ 
           
           & \tabitem Navigation through unknown intention and noisy perception.& \cite{hubmann2018automated} \\
           
           & \tabitem Inferring High-level motion intentions including turning and going straight. & \cite{song2016intention,shu2020autonomous} \\ 
           
    \hline
    \hline
    \end{tabular}
    \label{table:3}
  \end{table}
\end{center}

\subsubsection{Training in Continuous Action Space}
\label{section3c:2}
\begin{figure}[!ht]
    \centering
    \includegraphics[height= 8cm, width=8.4cm]{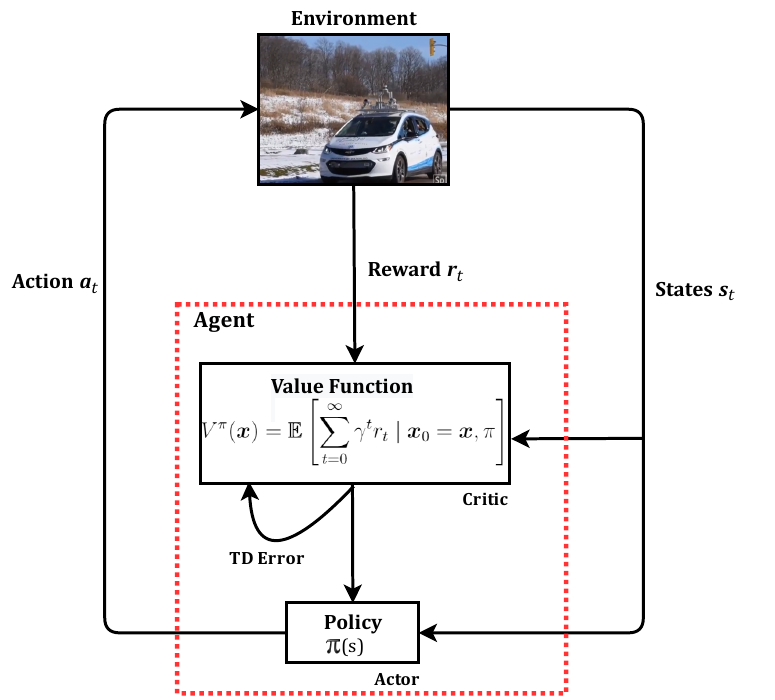}
    \caption [The LOF caption]{An illustrative sketch of Actor-Critic approaches. Actor-critic algorithms implement both the value-based approaches and the policy-based approaches. It comprises a couple of estimators: the actor network estimator which is based on Q-value, whereas the critic network utilizes the state-value function estimation.}
    
    \label{fig:actor}
\end{figure}

\textbf{DQNs and DDPG}. In real autonomous driving, a continuous action of the autonomous agent is required for safe and efficient navigational tasks. DQNs, which are mostly adopted in the reviewed decision-making schemes, are used to learn an optimal policy for safety-oriented decision-making in a discrete action space domain. However, adapting such schemes to continuous domains, i.e. autonomous driving, is considered challenging, and in some instances, sample inefficient. Practically speaking, DQNs determine an action that has the highest action-value through an iterative optimization process at every step in the continuous action. For complex multi-agent decision-making including urban intersections, we have high-dimensional continuous action spaces. Discretizing these spaces to use conventional DQN schemes is not always an effective idea due to the exponential number of action values which might lead to the \emph{Curse of Dimensionality}. Hence, to ensure convergence of the used model and capability, these continuous spaces must be handled in a robust way. Deep Deterministic Policy Gradient (DDPG) was adopted in \cite{qiao2018pomdp, li2021continuous} for generating continuous actions rather than discrete actions for driving in four-way unsignalized intersection settings. Xiong \emph{et al.} \cite{xiong2016combining} presented an integration between Deep Reinforcement Learning and safety-based continuous control for learning optimal policy for self-driving and collision avoidance applications. DDPG, which adopts the actor-critic concepts (see Fig.\ref{fig:actor}), is implemented to output the steering commands along with an Artificial Potential Field (APF) method for collision avoidance and path planning applications. As this integration proves its usefulness for learning collision-free driving policies at highways, integrating such high-level DRL schemes with control laws can be vital for solving continuous control problems within the framework of unsignalized intersections.

\textbf{TD3}. While the DDPG algorithm provides a solid foundation for handling continuous action spaces, it is often sensitive to hyperparameters and can fail due to overestimation of Q-values. To mitigate these issues, Twin Delayed DDPG (TD3) introduces key improvements that result in more robust and efficient learning \cite{fujimoto2018addressing}.
In contrast to conventional DDPG algorithms, which use a single Q-function to represent the quality or the utility of taking a specific action in a particular state, TD3 employs two Q-functions, \( Q_{\phi_1} \) and \( Q_{\phi_2} \), and utilizes the smaller of the two Q-values to form the targets in the Bellman error loss functions. This method, known as "Clipped Double-Q Learning," mitigates the issue of Q-value overestimation.
Further, in TD3, the policy and the associated target networks are updated less frequently than the Q-function. Specifically, for every two updates to the Q-function, a single policy update is performed. This "delayed" policy update reduces the volatility that typically arises in DDPG, enhancing the stability of the training process.
Finally, TD3 introduces "Target Policy Smoothing," a regularization technique that adds noise to the target action, which in effect smooths the Q-function along the changes in action space. This additional layer of complexity prevents the exploitation of erroneous sharp peaks in the Q-function, improving the robustness of the learned policy.
Xu \emph{et al.} \cite{9959664} showed TD3 to be particularly well-suited for the high-dimensional, continuous nature of autonomous driving at unsignalized intersections, outperforming DDPG in simulated unsignalized intersection traversal scenarios.

\textbf{MEDDPG}. Another common pitfall of conventional DDPG algorithms is their limited scope in exploring the state and action space, often resulting in local optima. This issue has been addressed by employing meta-learning techniques to adaptively learn exploration policies, as initially proposed by Xu \emph{et al.} \cite{xu2018learning}. This foundational work introduced the concept of using a meta-policy gradient to allow for more flexible and global exploration that is independent of the actor policy, thereby increasing the sample efficiency in DDPG training.
Building on this approach, Xu \emph{et al.} \cite{9959664} developed the Meta Exploration Deep Deterministic Policy Gradient (MEDDPG) algorithm specifically for the high-dimensional and complex nature of autonomous driving at unsignalized intersections. MEDDPG incorporates the adaptive exploration techniques from the original meta-policy gradient framework, but refines and optimizes them for the specific challenges posed by autonomous driving scenarios. By employing an independent meta-exploration policy that uses a stochastic policy gradient, MEDDPG aims to improve the learning rate of both the actor and critic networks, thereby facilitating more effective decision-making at unsignalized intersections. The algorithm outperforms conventional DDPG methods on unsignalized intersection traversal tasks by allowing for a more comprehensive exploration of the state and action space, thereby leading to faster convergence and more robust policies.

\textbf{T-TD3}. Traditional Markov Decision Processes (MDPs) operate on the premise that the future state is dependent solely on the current state and action, disregarding any past states. This assumption becomes questionable in the context of autonomous driving at unsignalized intersections. Traffic dynamics are inherently temporal, and past states often provide crucial information for optimal decision-making. Ignoring previous states thus appears as an unreasonable simplification for this application, given that the decision-making of an autonomous vehicle (AV) often depends on the temporal trends of its surrounding environment.
Building upon the work of DeepMind \cite{mnih2013playing}, Xu \emph{et al.} \cite{9959664} also developed a specialized approach for autonomous driving termed Time Twin Delayed Deep Deterministic Policy Gradient (T-TD3). This algorithm expands the traditional state space S to incorporate past states, thereby giving a richer context for decision-making.
This enhancement allows the model to be more aware of the temporal intricacies in the driving environment, thus overcoming the limitations inherent to traditional MDPs. In addition to this, T-TD3 employs Long Short-Term Memory (LSTM) cells to better account for time-series data, which is essential for understanding the motion trends of surrounding vehicles.
The use of LSTMs allows the model to dynamically adapt to the environment and predict actions based on the history of states, rather than just the current state. Consequently, the T-TD3 algorithm provides a more robust and real-time decision-making mechanism for autonomous driving at unsignalized intersections, marking a significant step forward compared to existing MDP-based methods.

\textbf{Soft-actor-critic (SAC)}. Recently, soft-actor-critic (SAC) algorithm has shown better performance in learning policies in continuous domains and stability characteristics than other deep deterministic algorithms including DDPG \cite{haarnoja2018soft}. For autonomous driving applications, SAC has demonstrated remarkable ability in learning optimal policies for overtaking and maneuvering at roundabouts \cite{song2021autonomous,liu2021improved}. Hence, applying soft-actor-critic (SAC) principles for learning traversing policies in complex driving scenarios can be a significant research avenue to be pursued.

\textbf{Safety, Efficiency, Comfort, Energy-saving}. Yuan \emph{et al.} \cite{yuan2023safe} investigate the applicability of various DRL algorithms (DDPG, PPO, TRPO) for autonomous driving at unsignalized roundabouts. Specifically, they design a comprehensive reward function that amalgamates aspects of safety, such as lane adherence and time-to-collision, with the efficiency of movement, comfort through smooth vehicular control, and energy consumption gauged by Vehicle Specific Power (VSP).
The authors adopt a 7-feature representation that encompasses both vehicle and environmental parameters and a hybrid action space that blends both discrete meta-actions for behavioral patterns and continuous actions for precise throttle and steering controls.
The empirical assessment of the DRL algorithms underlines the nuanced balance of the reward function, with the TRPO algorithm excelling in safety and efficiency, while PPO emerges as the superior choice for comfort and energy consumption. Moreover, the adaptability of the TRPO-trained model is validated through its commendable performance in additional driving scenarios like highway navigation and merging.

\subsubsection{Training in high-dimensional state-action space}

As mentioned earlier in \ref{section2d}, DRL is centered on performing iterative optimization processes to learn a policy for a specific task. However, the number of iterations grows exponentially as the state-action space becomes larger. One discernible shortcoming of adopting DQN and DDPG is that extensive training has to be performed in order to achieve optimal behaviour.

\textbf{Curriculum Learning}. To accelerate the training process, Curriculum Learning (CL) approaches can be employed \cite{bengio2009curriculum}. Qiao \emph{et al.} \cite{qiao2018automatically} utilizes Curriculum Learning for reducing the training time and improving the performance of the agent in unsignalized intersection approaching and one-dimensional crossing behavior. However, applying CL concepts for other more complex scenarios. i.e. two-dimensional left-turn was not investigated. 

Khaitan \emph{et al.} \cite{9981109} propose a state dropout-based curriculum for PPO to further improve agent performance. Their methodology introduces two distinct curricula aimed at overcoming the issue of suboptimal policies that often plague RL agents in unsignalized intersections by adopting an unconventional approach: unlike traditional curriculum learning strategies that transition from simpler tasks to more complex ones, Khaitan \emph{et al.} start with the full complexity of the task but provide the agent with additional, privileged information. In the first curriculum, training begins by providing the agent with complete future state information for surrounding vehicles for N time-steps. As the agent progresses through subsequent training phases, it gradually loses access to this future state information, compelling the agent to adapt and learn from increasingly incomplete data. A second curriculum takes a more dynamic approach: it augments the agent's action space with an additional "prediction" action. This allows the agent to choose when to discard future state information, thereby gaining a more nuanced understanding of its environment. Importantly, the reward structure is also adapted to incentivize the agent to drop more future state information as it becomes more competent, rewarding the agent with higher scores for predicting the behavior of surrounding vehicles without the aid of future state information (See II-B on Driver Intention Inference algorithms). Both curricula outperformed standard PPO and rule-based TTC methods at traversing unsignalized intersections.

\begin{figure}[!ht]
    \centering
    \includegraphics[height= 5.5cm, width=9.2cm]{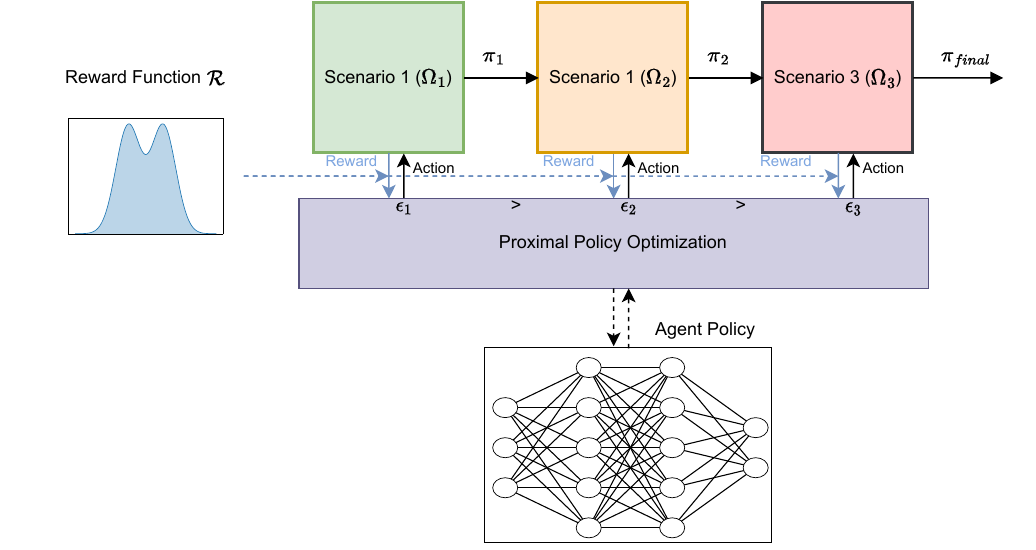}
    \caption [caption]{Curriculum learning pipeline}
    
    \label{fig:cl}
\end{figure}

Zengqi Peng \emph{et al.} \cite{peng2023curriculum} have developed the Curriculum Proximal Policy Optimization (CPPO) framework, enhancing the standard PPO algorithm for autonomous vehicles at unsignalized intersections (see Fig.\ref{fig:cl}). CPPO introduces stage-decaying clipping to balance exploration and exploitation during training, adapting its clipping parameter in stages for optimal policy convergence. Initially, a higher clipping parameter allows broad exploration, followed by a reduced parameter for refined policy search. Integrating curriculum learning, CPPO progressively increases training scenario complexity, improving the agent's adaptability and generalization. The framework utilizes a multi-objective reward function, varying rewards and penalties based on scenario complexity and traffic density, encouraging safer and more efficient driving strategies. Comparative simulations show CPPO's superior adaptability and faster training convergence than baseline PPO methods. However, its validation has been limited to simulations, with future research needed in sim-to-real transfer.

\textbf{RNN and LSTM}. In \cite{qiao2018pomdp}, Qiao \emph{et al.} proposed a DRL learning algorithm to traverse a four-way intersection with a two-way stop sign by taking into account the uncertainties that exist in the urban environment. This DRL algorithm is developed to utilize the preserved state-action values and the current LIDAR information along with the ego car's state information for designing the decision process. For efficient training in a high-dimensional space, a combination of LSTM and FC neural networks were designed to store the state-action pairs and generate continuous actions. Bouton \emph{et al.} proposed a DRL algorithm for navigating urban intersections using the scene decomposition method \cite{bouton2019safe} to improve training and to scale to a large number of agents. The decision-making under faulty perception is modeled as a POMDP. An extra state variable has been integrated to the global state vector to model the potential incoming traffic participant which is not present in the scene. A probabilistic model checker was adopted to compute the probability of reaching the goal safely for each state-action pair prior to learning a policy. Subsequently, a belief updater algorithm was developed to update the state's uncertainties. Given the prior belief value and the current observations, the algorithm can integrate the perception error into the planning theme. It uses an ensemble of 50 Recurrent Neural Networks (RNN) to store the observation history. The training process was done using a synthetic dataset generated from a simulation environment. These techniques, however, have not been evaluated in real-world driving scenarios, where convergence of the proposed models is not guaranteed due to the breadth of possible crossing behaviors or directions of agents at unsignalized intersections.

\textbf{Attention-based Schemes}. Traditional driving policies often struggle with high-dimensional state-action spaces and have limited ability to focus on the most pertinent features within a driving scenario. To overcome these challenges, Seong \emph{et al.} \cite{9564720} propose an attention-based driving policy that leverages deep reinforcement learning. The model's state representation comprises three components: a pseudo-scan of surrounding vehicles, a local trajectory related to the vehicle's path, and the vehicle's dynamic state. The action space is defined in terms of the vehicle's target velocity in a continuous space. Critically, the model integrates spatial and temporal attention mechanisms into its policy network, allowing it to focus on the most relevant spatial and temporal features in the driving environment. This focused attention makes the model's decisions both safe and efficient across a range of complex intersection scenarios and varying traffic densities. The model's performance is quantified through extensive experiments that show it outperforms baseline policies. Moreover, the attention-based design lends itself to interpretability, an often sought but rarely achieved attribute in deep learning models. Importantly, the authors validate the real-world applicability of their model by successfully transferring it to a full-scale vehicle system, demonstrating its robustness even in the presence of sensory noise and delayed responses.

\textbf{Graph Networks}. Graph Attention Networks (GATs) \cite{velickovic2018graph} are increasingly used for modeling environments in autonomous driving, especially at unsignalized intersections. The DQ-GAT \cite{9819830} method combines deep Q-learning with GATs and noisy networks for efficient exploration, automatically adjusting exploration noise.  It leverages noisy linear layers \cite{fortunato2019noisy} for balance in exploration and exploitation, supporting stable policy deployment. Asynchronous training using the Deep Q-Network (DQN) algorithm enhances training efficiency by generating diverse experiences simultaneously. The model uses semantic abstraction with Bird's Eye View (BEV) images for dimensionality reduction, providing a geometrically consistent perspective that aids in learning and generalization, but also presents a limitation in real-world driving scenarios where a Bird's Eye View (BEV) is not naturally available. The graph model captures interactions among traffic agents using a two-layer GAT with multi-head attention, representing each agent as a graph node. DQ-GAT outperforms other DRL methods in training comparisons, with visual analyses showing its effectiveness in real scenarios like complex intersections. Enhanced by policy visualizations, the model's transparency is evident. Its successful application on the openDD dataset \cite{breuer2020opendd} and high inference speed suitable for real-time applications underscore its effectiveness and practical utility.

\begin{figure}[!ht]
    \centering
    \includegraphics[height= 5.9cm, width=9cm]{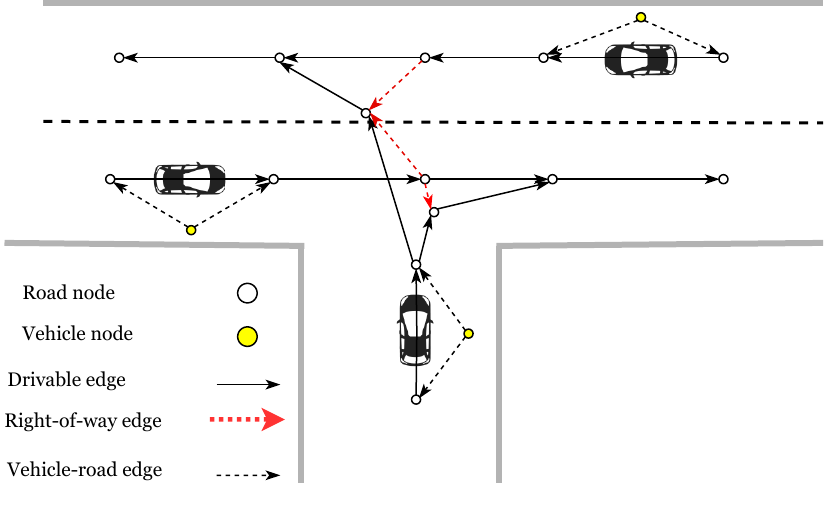}
    \caption [caption]{Example heterogeneous direct graph representation of road topology and vehicles used in \cite{10160762}}
    
    \label{fig:topo}
\end{figure}

In \cite{10160762}, Schier \emph{et al.} highlight the limitations of current graph-based approaches which fail to encompass the entire road network and overly depend on handcrafted features for vehicle-to-vehicle interaction modeling as shown in Fig.\ref{fig:topo}. To address these shortcomings, the authors propose a framework that captures the complexity of road networks and traffic participants within a heterogeneous directed graph. This representation can handle different elements (e.g., various types of vehicles, pedestrians, cyclists, traffic signs) and their distinct properties, thus capturing the complexity of the road network and its users, unlike traditional graphs that may not capture the full scope and rely on static, handcrafted features.
This graph is then adeptly transformed into a simpler vehicle graph with learnable edges, representing the routes connecting the vehicles. This allows for the reinforcement learning algorithm to operate on a simplified but effective representation of the environment, focusing on the dynamic interactions of vehicles while navigating the roads.
Their experimental validation in SUMO demonstrated a significant improvement in the performance of the proposed scheme with learnable edge features. This enhancement indicates a more effective representation of vehicle relations.

\textbf{GPT-based Approaches}. Liu \emph{et al.} \cite{liu2023mtd} propose their MTD-GPT model to harness the synergy of reinforcement learning (RL) and the transformative sequence modeling capabilities of Generative Pre-trained Transformers (GPT), setting a new stage for managing complex driving tasks.
They first develop single-task RL expert models using Proximal Policy Optimization (PPO) combined with an attention mechanism. These experts were trained to excel in specific driving tasks such as turning left, moving straight, or turning right. After training, the experts' decision-making, embodied in state-action-reward sequences, was captured as data.
This expert data served as a foundational guide for the MTD-GPT. The data was tokenized, akin to how text is prepared for natural language processing, to fit the GPT model's input format. The MTD-GPT was then trained offline in an autoregressive fashion to predict actions, drawing on this mixed multi-task dataset for learning.
The training of MTD-GPT involved embedding the tokenized state-action-reward sequences, adding positional encodings, and feeding them through a Transformer architecture. The self-attention mechanism within the Transformer was key to focusing on relevant parts of the input sequence for making decisions. The model was tuned to predict actions based on the input sequence, effectively learning from the collected expert demonstrations.
Upon evaluation, the MTD-GPT model demonstrated its robustness by performing on par with or even surpassing dedicated single-task RL models. This indicates that the model can abstract the multi-task decision-making problem into a sequence prediction task, providing a promising research direction.

\textbf{Learning from Demonstrators (LfD).} Learning from demonstrators has been introduced to enhance the learning capabilities, yielding a significant decrease in the total time of the training process by leveraging training sets with small demonstrations. Hester \emph{et al.} \cite{hester2017deep} introduced Deep Q-learning from demonstrations (DQfD) to significantly accelerate the training process through leveraging sets with small demonstrations. A prioritised replay mechanism was adopted for assessing the required data-sets ratio automatically. Nair \emph{et al.} \cite{nair2018overcoming} proposed a technique based on DDPG and Hindsight Experience Replay for enhancing the training policies while learning the optimal policy for solving complex tasks using RL. Although, the proposed work has one major limitation which is the sample efficiency, a significant speed-up of the training process was recorded. These works have led to other modifications of the training process of RL-based motion planning schemes. For instance, Hierarchical Reinforcement Learning (HRL) architecture was developed based on the inclusion of heuristic-based rules-enumeration policy to enhance agent's exploration for behavioral planning at intersections \cite{qiao2021behavior}.

\textcolor{black}{Incorporating Imitation Learning (IL) approaches can accelerate training and improve safety by reducing the reliance on random exploration \cite{zare2024survey}, which is particularly crucial in complex, multi-agent environments like unsignalized intersections \cite{}. However, IL alone may struggle with generalization \cite{cheng2024rethinking}, as it heavily depends on the quality and diversity of the expert demonstrations. On the other hand, RL-based approaches, while more autonomous and capable of handling novel scenarios, often require extensive interaction with the environment, making them computationally expensive and potentially unsafe during the early stages of training \cite{al2023self}. Despite these strengths, RL-based decision-making remains challenging for real-world applications, particularly in safety-critical environments such as unsignalized intersections. One significant issue is the sample inefficiency of RL, where learning robust policies requires a massive amount of data and interactions. This is compounded by the difficulty in designing appropriate reward functions that align with safe driving behavior, as poorly designed rewards can lead to unintended or unsafe actions \cite{zhu2021survey}. Furthermore, real-world deployment of RL-based systems is hindered by difficulties in simulating edge cases and handling unexpected situations, which are crucial for ensuring safety in autonomous driving \cite{le2022survey}.
\\
By integrating IL techniques with RL, the agent can benefit from expert knowledge while still exploring and adapting to new situations, addressing some of the key challenges in RL. This combination can also improve the safety and efficiency of training in high-risk environments like unsignalized intersections. However, this integration introduces challenges, such as the potential for overfitting to expert demonstrations or difficulties in determining when to switch from imitation to reinforcement learning. Addressing these issues will be crucial for developing decision-making architectures that are both effective and safe in real-world unsignalized intersections. Recently, Huang et al. \cite{huang2022efficient} developed an integration between imitative expert priors from expert demonstrations for learning driving policies in urban environments. In this approach, the priors are learned using IL and an uncertainty quantification method, which helps balance exploration and exploitation during training. This integration highlights the complementary nature of IL and RL: while IL provides a more stable and safer foundation by leveraging expert demonstrations, RL enables further refinement through exploration, allowing the agent to adapt to unseen scenarios.}

In brief, Table \ref{table:4} epitomizes the limitations of the most relevant research works on decision-making using reinforcement learning-based schemes.

\begin{center}
  \begin{table*}[t]
    \caption{Overview of the reviewed Reinforcement learning-based decision-making schemes at unsignalized intersection}
    \centering
    {\fontsize{5.91}{6.65}\selectfont 
    \begin{tabular}{ c|p{2cm}|p{2.5cm}|p{2cm}|p{4cm}|p{4.5cm}}
     \hline \hline
      \textbf{Ref.} & \textbf{Intersection Type} & \textbf{Method} & \textbf{Data Collected} & \textbf{Remarks} & \textbf{Limitations}\\ \hline
      
      \cite{qiao2018automatically} & 4-way Unsignalized (stop-sign) & DRL using automatic generation of curriculum for training enhancement   & Simulation-based & \tabitem A more realistic 4-way intersection driving scenario is proposed where vehicles are programmed not to yield to ego vehicle if it is in their FOV & \tabitem Environmental uncertainties, which cause errors in perception were not considered while collecting Observations from simulated sensor (LIDAR + Cameras) \\ \hline

      \cite{isele2018navigating} & T-junction & DQN & SUMO simulator \cite{krajzewicz2012recent}& \tabitem  A creeping behavior upon reaching the intersection is learnt where the agent must perform an exploratory action to better comprehend the environment& \tabitem A god-view state space is used to describe the motion of vehicles at intersection (Not true for real-life driving) \\ \hline

      \cite{li2019urban}& Multi-lane 4-way intersections and roundabouts. & Multi-objective RL (Thresholded lexicographic Q-learning) &  Collected Via SUMO simulator & \tabitem Learning safe crossing with the presence of faulty perception and occlusion \newline \tabitem The trained policy is scalable across a range of urban roads with different shapes \newline \tabitem Learning human behavior of looking at vehicles at area of interest & \tabitem A full knowledge of other vehicles based on a bird's eye view representation of state space (not realistic for real-world) \newline\tabitem Not tested in real-world environments \\ \hline
       
      \cite{ qiao2018pomdp} & 4-way Unsignalized (stop-sign)& RL-based approach using hierarchical option & Collected Via SUMO simulator & \tabitem Learning an optimal policy for robust traversing under environmental uncertainties \newline\tabitem Results shows superiority over the rule-based techniques and classical approaches &\tabitem No guarantees for possible scalability at more complex intersections with multi-lanes\newline\tabitem Not tested in real-world environments \\ \hline
       
      \cite{bouton2019safe} & T-junction & Integration of model-checker and RL & Simulation-based & \tabitem Learning safe crossing with the presence of faulty perception and occlusion & \tabitem The proposed method was not validated through real testing to show the validity of the simulated POMDP-based simulated values of the perception errors\\ \hline

      \cite{qiao2021behavior} & Multi-lane 4-way intersection & Hierarchical reinforcement learning with hybrid reward mechanism & MSC’s VIRES VTD (Virtual Test Drive) simulator & \tabitem Better convergence capabilities  and sample-efficiency compared to the classical RL Methods& \tabitem Focus on mimicking human driving in limited go-straight and left-turn maneuvers \newline\tabitem Not tested in real-world environments \\ \hline
       
      \cite{kamran2020risk} & Multi-lane 4-way intersection.& DQN & Collected Via CARLA Simulator \cite{dosovitskiy2017carla}) &\tabitem DQN shows less overcautious behavior under limited sensor range and faulty perception compared to the rule-based algorithms& \tabitem DQN is utilized for learning the driving policy. However, DQN is restricted to the discrete action domain \newline\tabitem Not tested in real-life environments\\ \hline

      \cite{liu2020decision2}& 4-way& DQL and DDQL & Simulation-based & \tabitem The proposed results show safe and repeatable Left-turn maneuver is learnt where the collision rate is significantly reduced & \tabitem Training based on simulated sensor observations. \newline \tabitem The proposed scheme is restricted to discrete action space. \\ \hline
              
      \cite{hoel2020reinforcement} & 4-way & Bayesian RL-based scheme using an ensemble of  NN  with  Randomized  Prior Functions  (RPF) & Simulation-based & \tabitem The Uncertainty of the RL agent’s actions is estimated.& \tabitem Lacks real-world testing \newline \tabitem Assumptions related to the formulated decision-making problem have been made, i.e. the environment is assumed to be fully observable (MDP) \\ \hline

      \cite{bouton2019reinforcement} & T-junction & RL with stochastic guarantees & Simulation-based & \tabitem Traversing with safety guarantees & \tabitem The proposed scheme deals with discrete action space only \newline \tabitem Assumptions made for the vehicle and the pedestrians motion\\\hline
      
      \cite{shu2022driving} & 4-way & DQL and DDQ & Simulation-based & \tabitem RL-enabled control framework is built using transfer rules & \tabitem RL scheme deals with discrete action space only. \tabitem The proposed geometric controller does not represent actual vehicle constraints, e.g. max steering rate. \\ \hline

      \cite{9981109} & Multi-lane 4-way intersection \& Unsignalized T-junction & State Dropout-Based Curriculum Reinforcement Learning & Simulation-based (CommonRoad) & \tabitem State dropout-based curriculum learning approach with PPO & \tabitem RL scheme deals with discrete action space only. \newline \tabitem Lacks real-world testing \\ \hline

      \cite{peng2023curriculum} & Multi-lane 4-way intersection & Curriculum Proximal Policy Optimization & Simulation-based & \tabitem Rapid policy search and optimal convergence through adjustable clipping in PPO \newline \tabitem Improved generalization and speed via stage-based curriculum learning. & \tabitem RL scheme deals with discrete action space only. \newline \tabitem Lacks real-world testing \\ \hline

      \cite{9959664} & Multi-lane 4-way intersection & DDPG, MEDDPG, TD3 \& T-TD3 & Simulation-based & \tabitem Novel time twin delayed DDPG algorithm to mitigate overestimation bias in value approximation and improve training stability \newline \tabitem Novel Meta Exploration Deep Deterministic Policy Gradient (MEDDPG) algorithm combines meta-learning principles with exploration strategies to allow for more global exploration & \tabitem Lacks real-world testing \newline \tabitem Assumptions made about other vehicles' motion (fixed speed) \\ \hline

      \cite{9819830} & Multi-lane 4-way intersection (unprotected left turn), Unsignalized T-junction \& Unsignalized roundabout & DQ-GAT: DQN and GATs & Simulation-based (CARLA) and real-world dataset openDD & \tabitem GATs to model interactions in dynamic traffic, and DQN for learning scheme & \tabitem RL scheme deals with discrete action space only. \newline \tabitem Proposed scheme relies on BEV images \\ \hline

      \cite{10160762} & Multi-lane 4-way intersection, Skewed Multi-lane 4-way intersection, Unsignalized Merge Junction, \& Unsignalized roundabout & High-Level Heterogeneous Graph Representations & Simulation-based (SUMO) & \tabitem Use of heterogeneous directed graph to model full complexity of driving scenarios at unsignalized intersections \newline \tabitem Transformation into a simpler vehicle graph with learnable edges for better integration with DRL scheme & \tabitem RL scheme deals with discrete action space only. \newline \tabitem Lacks real-world testing \\ \hline

      \cite{9564720} & Multi-lane {3, 4, 5}-way intersection, Unsignalized roundabout & Attention-based Deep Reinforcement Learning & Simulation-based (CARLA) \& real-world deployment & \tabitem Use of spatial and temporal attention focus mechanism & \tabitem The developed DRL technique is agnostic to motion planning and control layers  \\ \hline

      \cite{liu2023mtd} & Multi-lane 4-way intersection & MTD-GPT & Simulation-based (OpenAI Gym) & \tabitem GPT-based model based on single-task DRL expert models (PPO) & \tabitem Lacks real-world testing \\ \hline

      \cite{9922164} & Multi-lane 4-way intersection & Safe and Rule-Aware Deep Reinforcement Learning & Simulation (SUMO \& CARLA) & \tabitem DRL framework w/ traffic rule monitor to ensure compliance with intersection right-of-way rules \newline \tabitem RSS-based safety checker used to identify and react to unsafe situations & \tabitem RL scheme deals with discrete action space only. \newline \tabitem Lacks real-world testing \\ \hline

    \hline
    \hline
    \end{tabular}
    }
    \label{table:4}
  \end{table*}
\end{center}

\section{Discussion and Research Directions}

\begin{center}
  \begin{table}[t]
    \caption{Performance evaluations}
    \centering
    {\fontsize{8}{8}\selectfont 
    \begin{tabular}{ c|p{1.8cm}|p{1.5cm}|p{1cm}|p{1cm}}
     \hline \hline
      \textbf{Ref.} & \textbf{Intersection Type} & \textbf{Maneuver} & \textbf{Success rate (\%)} & \textbf{Collision Rate(\%)} \\ \hline

      \cite{qiao2018automatically} & 4-way intersection &  Straight & 82.1  & 13.5 \\ \hline

      \cite{qiao2018pomdp} & 4-way intersection & L. Turn \newline R. Turn \newline Straight & 97.3 \newline 99.8 \newline 98.3 & 2.6 \newline  0.2 \newline  1.7 \\ \hline

      \cite{li2019urban} & Multi-lane intersections &  Turning \newline Yielding &  N/A &  3.5 \\ \hline

      \cite{9564720} & Multi-lane intersections &  L. Turn \newline R. Turn \newline  Straight &  87 \newline 97 \newline  92 &  5 \newline  2 \newline  3 \\ \hline

      \cite{qiao2021behavior} & 4-way intersection & Straight &  95.6 & 4.2 \\ \hline

      \cite{9981109} & 4-way intersection & L. Turn &  93.83 & N/A \newline N/A\\ \hline

      \cite{9819830} & {3, 4, 5}-way intersection & L. Turn  & 96.67 &  N/A \\ \hline

      \cite{9922164} & 4-way intersection & Straight & 98.6 & N/A \\ \hline

      \cite{liu2022improved} & Roundabout & L. Turn & 93 & 5 \\ \hline

      \cite{shu2022driving} & 4-way intersection &  L. Turn \newline R. Turn \newline  Straight & 90 \newline 92.5 \newline  94 &  N/A \\ \hline

      \cite{10160762} & 4-way intersection &  L. Turn & 90 & 10 \\ \hline

      \cite{peng2023curriculum} & 4-way intersection & L. Turn & 90.5 & N/A \\ \hline

      \cite{cai2023rule} & 4-way intersection & L. Turn & 94.7 & N/A \\ \hline

      \cite{al2023self} & 4-way intersection & L. Turn & 97.8 & 1 \\ \hline

    \hline
    \hline
    \end{tabular}
    }
    \label{table:5}
  \end{table}
\end{center}

Based on our thorough investigation, we conclude that the state-of-art decision-making schemes focused on the high-level decision making layers, i.e high-level reasoning for behavioral path planning, neglecting other low-level layers proposed earlier, including low-level motion planning and control \cite{daoud2022simultaneous, dempster2023real}. Furthermore, implementation and testing in real-world driving environments is not investigated. In practice, convergence of the RL-models in simulation-based environments does not necessarily ensure generalizability in real-life scenarios due to the domains mismatch. Real-world observations differ in terms of the associated noise sequences and vehicle dynamics response. Therefore, in this section, we suggest avenues for research built upon these insights with the aim of progressing the field of study.

\subsection{Motion planning and Low-level control integration} 
\textbf{Model Predictive Control (MPC) Employment.} Numerous research papers have addressed the motion planning problem and control at urban unsignalized intersections using MPC principles. For instance, Hu \emph{et al.} \cite{hu2021event}, proposed an event-triggered model predictive adaptive dynamic programming technique for motion planning at urban intersections. The method takes urban speed, vehicle kinematics and road constraints into consideration while solving multi-objective optimization problem. Recently, in \cite{bautista2022autonomous}, an integrated control technique combines reinforcement learning and model predictive control is proposed for motion planning at unsignalized T-intersections. The integration envelops two independent control systems: one that involves nominal RL longitudinal control and path selection, and another for reactive MPC longitudinal cruise control and lateral path tracking. These systems are capable of independently managing the vehicle's lateral and longitudinal dynamics. This coupled RL/MPC architecture serves as a backup mechanism to enhance navigation safety. The two algorithms run concurrently, and a discrete selector focused on safety considerations determines the control output. Nevertheless, this proposed scheme focuses on the motion planning an control without considering the behavioral path planning layer. Additionally, the proposed solution is computationally intensive due to the independent functioning of the MPC model and reinforcement learning-based control mechanisms, along with a supervisory system responsible for output selection. Another integration has been done for intersection-management applications, where centralized reference signals being distributed to the intersections agents via V2V communication \cite{wang2021distributed, hamouda2021multi}. Hamouda \emph{et al.}, developed an integration between high-level decision-making layers and low-level MPC-based motion planing layer has been proposed for learning supervisory intersection-management policy in connected driving fashion. Nonetheless, for this solution to be completely workable, it is crucial that vehicular communication and connected vehicle technologies are widely distributed—a situation that has not yet been realized.

\begin{figure}[!ht]
    \centering
    \includegraphics[height= 8cm, width=9cm]{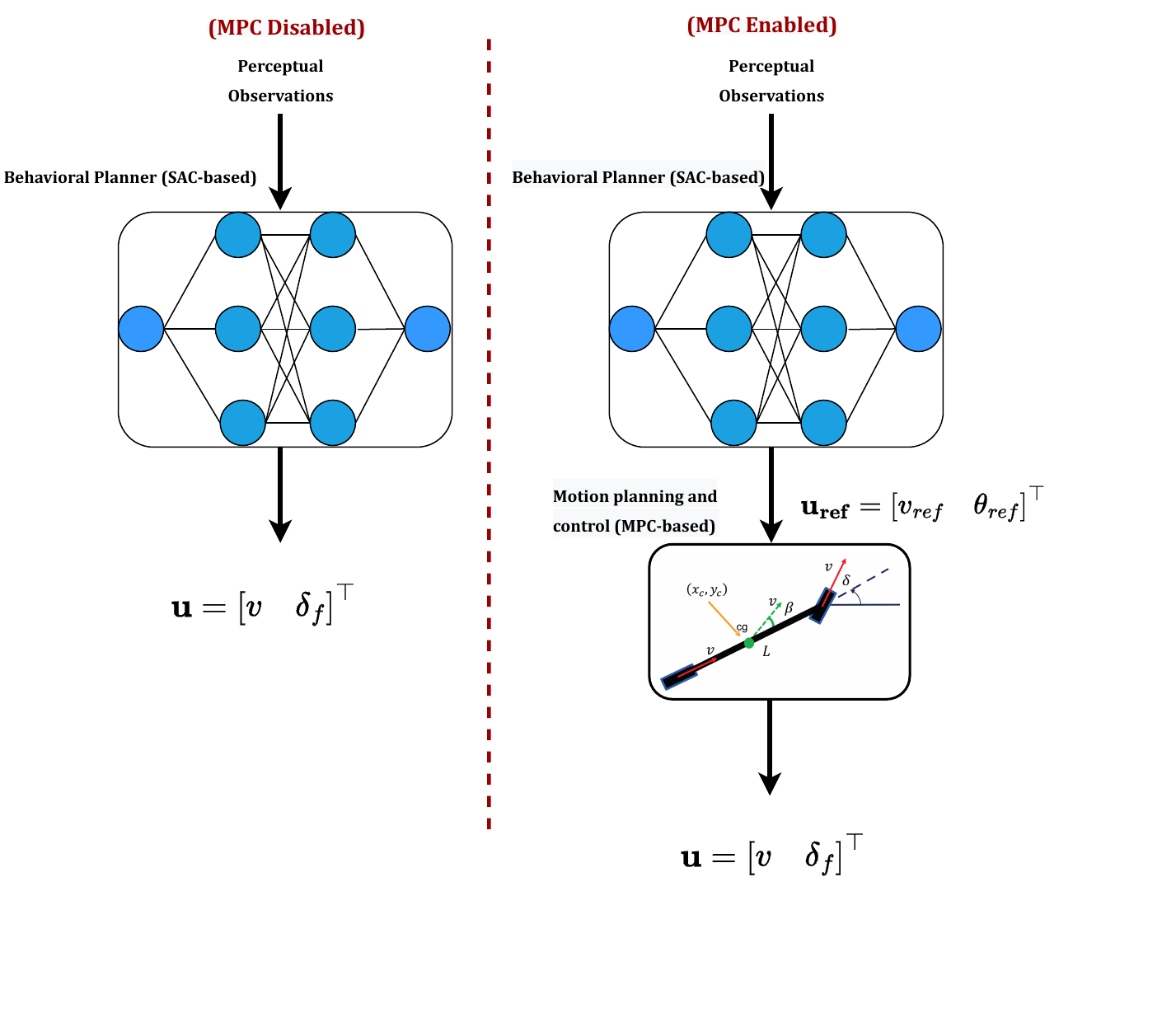}
    \caption [The LOF caption]{An illustrative sketch on the integration of Model Predictive Control with the SAC-based behavioral planning layer. When MPC is not integrated, the decision-maker (agent) receives perceptual observations from the intersection driving environment and maps them directly into throttle ${v}$ and steering ${\theta}$ commands executed by the environment. As a standalone RL setting, the traversing policy is trained through these interactions with the driving environment to provide actions to maximize the future rewards, regardless whether they are feasible or not. On the other hand, when the MPC is enabled, the policy is trained with the SAC algorithm to output reference velocity ${v}_{ref}$ and heading signals ${\theta}_{ref}$. The motion planning layer takes these reference signals as inputs to the two-dimensional tracking control problem, solving the formulated optimization problem while accounting for real-world constraints related to vehicle dynamics, urban traffic rules, and ride comfort. The optimized, feasible, control inputs are then produced to drive the vehicle’s physical model in the simulated driving environment.}
    
    \label{fig:1.5}
\end{figure}

In practical terms, achieving precise decision-making in urban autonomous driving necessitates the integration of motion planning and low-level control layers that consider vehicle dynamics with the RL-based behavioral planners. This integration is essential to ensure that the RL-based behavioral planner actions are feasible. Therefore, incorporating the motion planning layer while learning intersection-traversal policies would ensure from feasible actions and with high fidelity, taking into account lateral and longitudinal dynamics. To illustrate the importance of the proposed research direction, we developed recently a novel hierarchical reinforcement learning-based decision-making architecture for learning left-turn policies at unsignalized intersections with feasibility guarantees \cite{al2023self}. This hierarchical architecture is comprised of two distinct layers; a high-level learning-based behavioral planning layer which adopts soft actor-critic (SAC) principles to learn non-conservative, yet safe, driving behaviors and a low-level motion planning layer that uses Model Predictive Control (MPC) framework to ensure feasibility of the two-dimensional left-turn maneuver. The high-level layer is responsible for generating reference signals of velocity and yaw angles for the ego vehicle taking into account safety and collision avoidance behaviors with the intersection vehicles, whereas the low-level motion planning layer solves an optimization problem to track these reference commands taking into account vehicle dynamic constraints and ride comfort. We conducted several simulation experiments to demonstrate the significance of the proposed decision-making hierarchy. We observed that when utilizing MPC (refer to Fig. \ref{fig:1.5}), the policy begins to converge after 200k training steps, achieving a high rate of successful traversals. In contrast, the standalone SAC, when MPC disabled, converges at around 500k training steps which demonstrates significant sample efficiency. This difference can be attributed to the optimization of control inputs by the MPC, which already accounts for real-life driving constraints when interacting with the environment. 

We evaluated the suggested decision-making approach within simulated environments and performed a comparative analysis against alternative model-free Reinforcement Learning (RL) baseline methods. Our findings demonstrate that the proposed integrated approach, which combines SAC-based planning with MPC-based motion control, surpasses these baseline methods, including TD3 and PPO, in terms of training efficiency and navigation performance. Furthermore, the training results highlight the effectiveness of our method in terms of performance and sample efficiency. The testing results showcase the efficiency and safety of the learned left-turn behaviors, with a success rate of 97.8\% over 1000 testing episodes. Notably, despite the inherent challenges of navigating a complex two-dimensional left-turn intersection, the integration of the SAC-based behavioral path planning layer with the MPC-based motion planning layer leads to faster convergence and a higher success rate compared to existing literature (see Table \ref{table:5}).

As we emphasized the significance of hierarchical decision-making, which integrates decision-making layers for learning traversing policies in complex multi-agent environments, such principles can be applied to tackle the challenges posed by more intricate unsignalized intersection environments, characterized by occlusions and environmental obstructions that impede the attainment of accurate perception. Additionally, there is potential for enhancing the model's accuracy and navigation capabilities in the context of intersections with diverse shapes and geometries.

\begin{figure}[t]
    \centering
    \includegraphics[height=8.1cm, width=9cm]{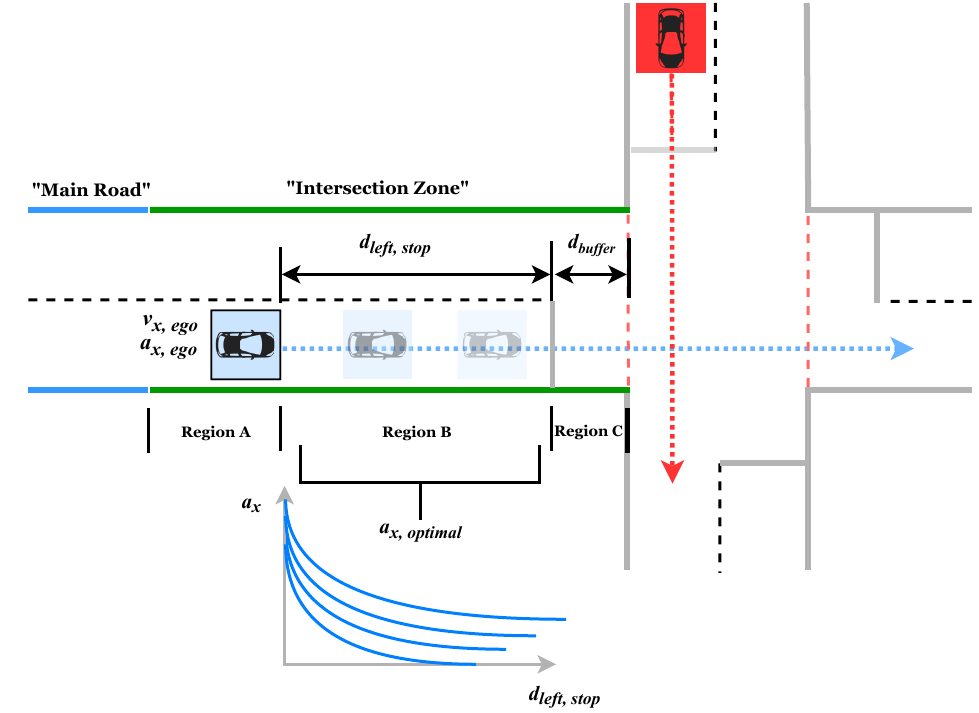}
    \caption{An illustrative sketch of the intersection-approach phase scheme. As shown, the vehicle enters \textbf{Region A} with the standard speed $V_{x,ego}$ of (40–50 km/h). In \textbf{Region B}, the vehicle is assumed to start decelerating with rate $a_{x,ego}$ to reach the stop-line. \textbf{Region C} represents the safety buffer $d_{buffer}$.}
    \label{fig:5.6}
\end{figure}

\textbf{Incorporating High-fidelity Vehicle Dynamic Models.} In connection with autonomous driving problem in adverse weather environments \cite{heidecker2021application,pitropov2021canadian,zhang2023perception}, incorporating high-fidelity vehicle dynamic models is also critical for longitudinal and lateral motion planning. For instance, learning safety-oriented policies for intersection-approaching behavior at unsignalized intersection, where braking is applied to decelerate smoothly for precise stopping, is a prerequisite condition for safe intersection navigation. Hence, learning an optimal deceleration profile (curve) $a_{x,optimal}$, which ensures a comfortable ride while remaining efficient, requires an inclusion of longitudinal vehicle dynamic models and braking performance which is coupled with the road surface conditions represented by friction coefficients (see rough curves in Fig. \ref{fig:5.6}).

\subsection{Real-world experimental validation} 

As can be seen from Table, \ref{table:4}, most of the reviewed schemes have been tested in simulation-based environments. This can be valid, as RL techniques require collecting a large amount of real-world based training data which would be costly in terms of effort and time. Practically speaking, simulated observations, which are streamed from modelled sensors, have different data distributions compared to real data which may lead to failure in generalization on (unseen) real data \cite{kiran2020deep}. 
This difference between simulated and real data distributions, such as inaccuracies in synthetic image generation or in vehicle dynamics, has been coined the "reality gap" (RG) \cite{10242366, 9606868}. It is known that agents trained in simulation transfer poorly to real-world environments without explicit regard for RG \cite{osinski2020simulation}. 
To rectify this, sim-to-real transfer learning techniques have been introduced to further promote training RL approaches in real environments \cite{pan2017virtual}. 

This survey highlights some proposed techniques which have been validated in real-world scenarios, and others which the authors believe to be promising theoretically or in other areas of robotics but require experimental validation in real-world scenarios using real-sized vehicles. Of those with validated results, we introduce Domain Randomization (DR) and Domain Adaptation (DA). While not tested outside of simulation, Adversarial RL techniques demonstrate improved robustness to environmental perturbations.

\textbf{Domain Randomization.} In utilizing DR methods, the agent is exposed to stochastic perturbations in the environment in simulation to improve robustness in real-world deployment \cite{peng2018sim, 8202133}. Inspired by H$^\infty$ control, such perturbations aim to provide robust control under (non) parametric uncertainty \cite{BHATTACHARYYA201745}, and allow the agent to be less sensitive to environment parameter perturbations. 
Kontes \emph{et al.} utilize domain randomization to design a high-speed collision avoidance controller for autonomous cars which, utilizing DR achieves near-perfect collision avoidance performance across all environment parameters than its peers trained using specific environment parameters \cite{9294396}. This suggests that utilizing DR to vary intersection angle and approaching car velocities may provide increased robustness of a proposed model traversing unsignalized intersections. 
Pouyanfar \emph{et al.} also demonstrate that DR provides resilience to sensor noise and sub-optimal operating conditions. \cite{Pouyanfar_2019_CVPR_Workshops} We note that this is of particular importance to driving at unsignalized intersections due to the variety of sensor occlusion patterns which may occur.
In addition, Amini \emph{et al.} \cite{amini2020learning} introduced a training engine for transfer learning of end-to-end autonomous driving policies using sparsely-sampled trajectories from human drivers.  Utilizing these trajectories has yielded robustness in performing real driving tests in unseen complex and near-crash environments. Using CARLA, the performance of the proposed method has been evaluated in comparison with the DA and DR approaches. The results exhibited superiority of the proposed approach over the conventional transfer learning approaches in terms of recovery from hazardous near-crash situation. 

\begin{figure}[!ht]
    \centering
    \begin{subfigure}{0.2\textwidth}
        \includegraphics[width=\textwidth]{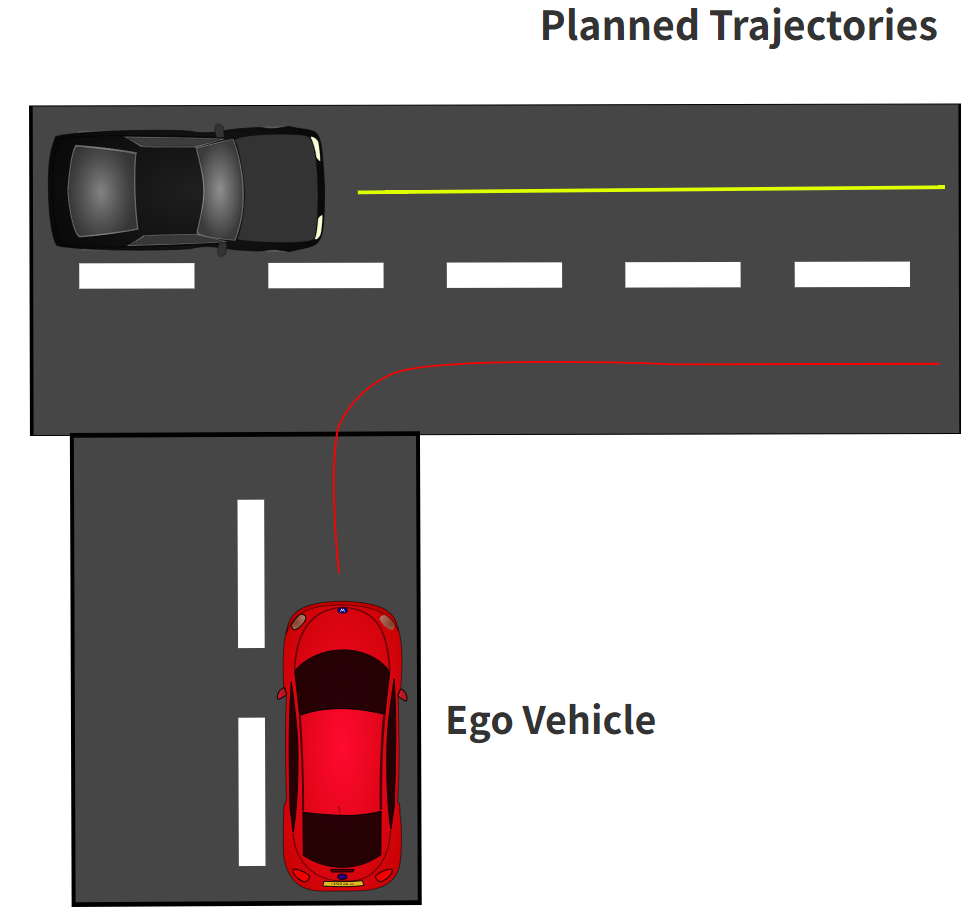}
        \caption{Simulation setup of turning at an unsignalized intersection}
        \label{fig:simsetup}
    \end{subfigure}
    \hfill
    \begin{subfigure}{0.2\textwidth}
        \includegraphics[width=\textwidth]{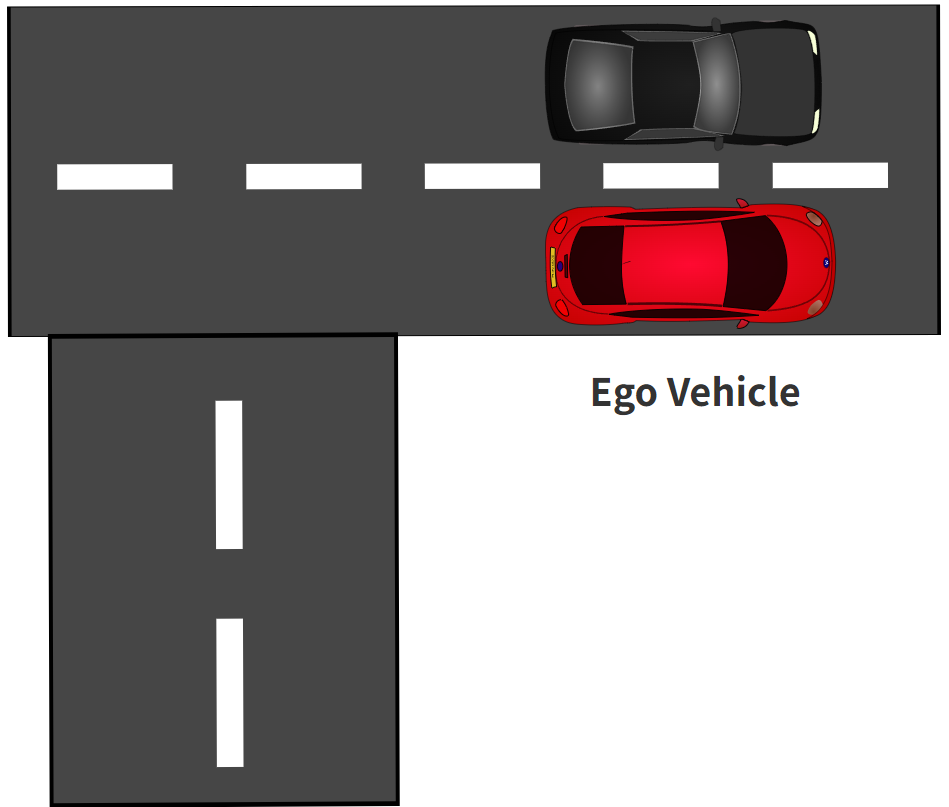}
        \caption{Simulation evolution of baseline RL models successfully turning in simulation}
        \label{fig:simnext}
    \end{subfigure}
    \hfill
    \begin{subfigure}{0.2\textwidth}
        \includegraphics[width=\textwidth]{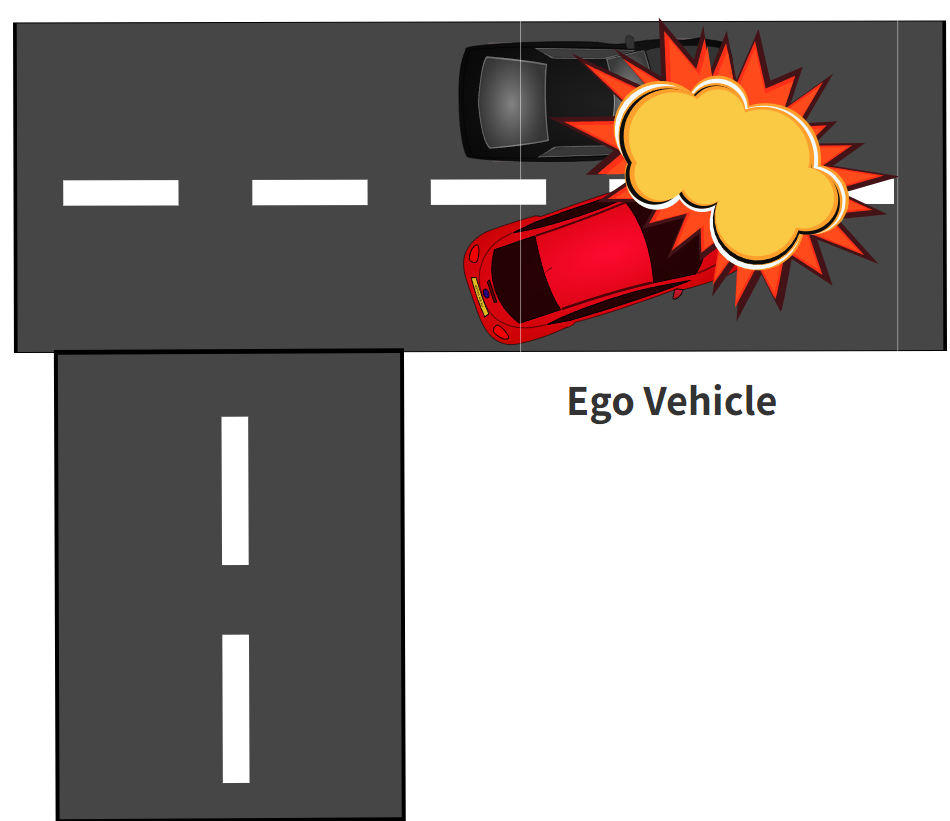}
        \caption{Irrecoverable adversarial perturbation causing crash}
        \label{fig:perturb_crash}
    \end{subfigure}
    \hfill
    \begin{subfigure}{0.2\textwidth}
        \includegraphics[width=\textwidth]{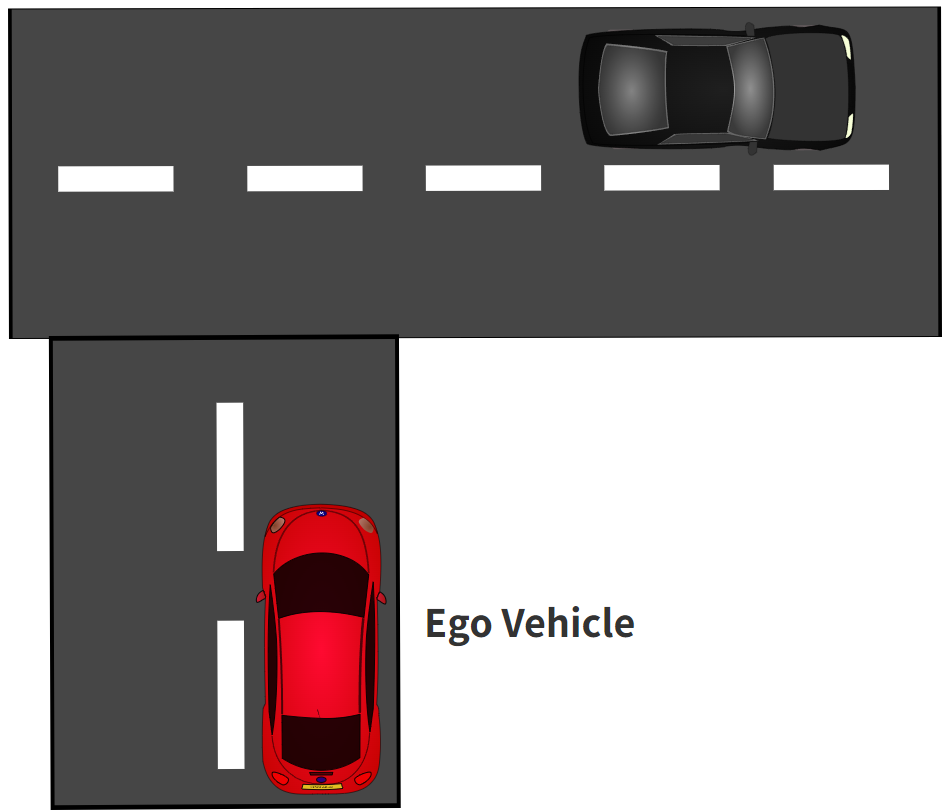}
        \caption{Risk-adverse algorithm learns to delay crossing to allow for spacing between cars before crossing}
        \label{fig:RARL_wait1}
    \end{subfigure}
    \hfill
    \begin{subfigure}{0.2\textwidth}
        \includegraphics[width=\textwidth]{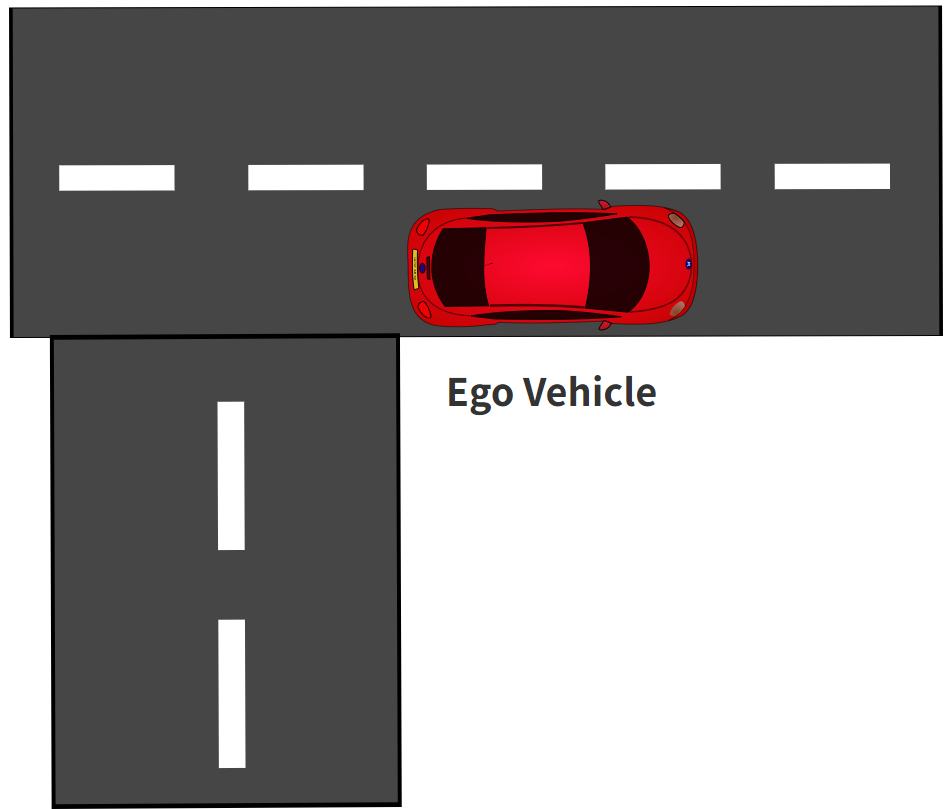}
        \caption{RARL agent turns only after first vehicle is a safe distance away}
        \label{fig:RARL_wait2}
    \end{subfigure}
    \hfill
    \begin{subfigure}{0.2\textwidth}
        \includegraphics[width=\textwidth]{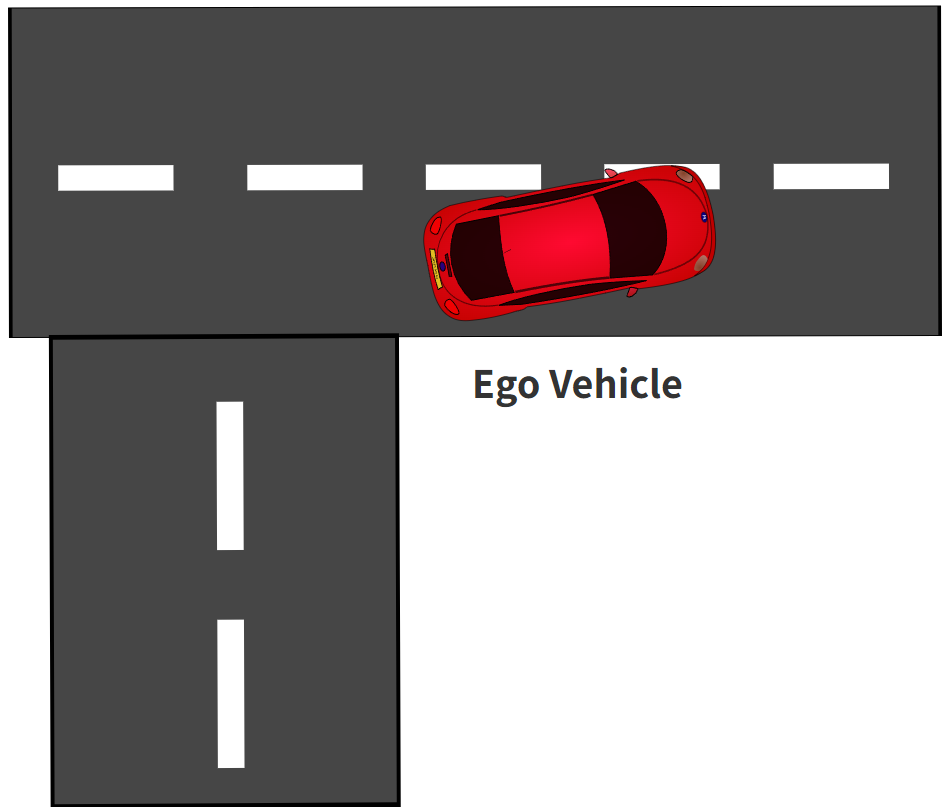}
        \caption{An adversarial perturbation causing a lane invasion, but because the ego vehicle is no longer adjacent to another vehicle this perturbation is recoverable}
        \label{fig:perturb_recover}
    \end{subfigure}
    \hfill
    
    \caption [The RARL caption]{Demonstration of a risk-adverse algorithm learning to delay rewards slightly in favour of less risky driving behaviour. The agent learns to avoid system states which are easily perturbed to a state of catastrophic failure, and prefers safer states where perturbed states are still safely recoverable from.}
    
    \label{fig:1.6}
\end{figure}

\textbf{Adversarial RL.} Taking inspiration from GANs, one may even adversarially perturb the environment as to mislead and destabilize the agent. Training an agent and adversary in this manner is similarly inspired by H$^\infty$ control methods and is known as Robust Adversarial Reinforcement Learning (RARL) \cite{pmlr-v70-pinto17a}. 
Presented with the task of merging onto highway lanes, a constrained adversarial RL policy consistently outperforms baselines in terms of speed and collision rates \cite{9994638}.
As a further improvement, rather than optimizing for the highest expected reward it is better to reduce the worst-case outcome in safety oriented applications. A limitation of RARL in this consideration is that only the expected reward is optimized without a modeling of risk.
To address this, Pan \emph{et al.} propose a risk-adverse RARL scheme where an adversarial agent explores states with high variance in value function to force the agent into risky scenarios \cite{8794293}. The resulting agent experiences substantially fewer collisions and safety-adverse events compared to previous baselines. This risk-adverse behavior, outlined graphically in figure \ref{fig:1.6} is exceptionally notable in autonomous driving where autonomous actors are expected to error at rates well below human errors. Furthermore, autonomous drivers must act in a way humans find agreeable from a high-level decision making standpoint. As such, risk-adverse algorithms are necessary to avoid what may be perceived as unnecessary risk and endangerment. As seen in figure \ref{fig:simnext}, baseline algorithms will favour riskier driving habits due to reward discounting. With the introduction of RARL schemes, the trained agent perform more inline with human behaviours and decision-making as seen in figures \ref{fig:RARL_wait1} and \ref{fig:RARL_wait2}. In addition, risk-adversity is especially important at unsignalized intersections to prevent trained models from crossing or turning when the guarantee of safety is low with high occlusion of passing cars. As previously noted, while the above RARL applications in AD indeed demonstrate improved robustness in environmental perturbations, further experimentation is required to validate the transfer into the real world.

\textbf{Domain Adaptation.} Domain Adaptation (DA) techniques have also been proposed to enhance the generalization capabilities of ML-based models on a target domain. Feature-level DA methods are designed to learn domain-invariant features which cannot discriminate between the source and the target domains, whereas pixel-level DA techniques focuses on shaping images from the source domain to be analogous to the target domain’s images using Generative Adversarial Networks (GANs) \cite{bousmalis2017unsupervised, hu2022sim}. Ganin \emph{et al.} \cite{ganin2016domain} describe a domain-adversarial training of neural networks for Feature-level domain adaptation. This model is based on features that are discriminative for the central learning process, but indiscriminative with the translation between these domains. In \cite{bewley2019learning}, an end-to end (i.e., perception to control) transfer learning using image-to-image translation for domain transfer was applied for autonomous driving. Although the lane following driving policy was learnt from the simulation domain with control labels, the model was able to provide control from real images due to the shared latent space between the two domains.

An inherent limitation of the autonomous driving sector is the cost associated with any real-world experimental testing such as vehicle and sensor replacement. Especially with tasks associated with non-negligible and unavoidable catastrophic failure rates such as high speed obstacle avoidance, there is often little incentive to test safety critical applications in a real world setting to experimentally gauge the transfer of learning from simulation to the real world.
However, there have been many promising sim-to-real transfer techniques proposed and tested in the real-world. Indeed, many others improving on previous techniques have not been validated. The authors hope that new sim-to-real techniques can be inspired by the experimental design of these techniques and demonstrate robustness extending to real-world applications.
In short, validating the RL approaches in real-world driving settings is an active area of research. Inspired by the presented sim-to-real techniques which prove its robustness in learning optimal policies for end-to-end autonomous driving, the real-world experimental validation of the simulation-based decision-making approaches would be further facilitated by creating real-life intersection driving scenarios.

\section{Conclusion}

 Unsignalized intersections are safety-critical areas in urban environments due to the complex driving behavior and the lack of traffic control signals. Consequently, developing robust decision-making and motion-planning for these multi-agent environments is highly intractable due to the complexities associated with the partially observable multi-agent driving environment and the environmental uncertainties. With the resurgence of deep learning, modern RL techniques have been utilized to handle such problems with a large space of observations to learn safe driving policies.  
 
 This survey reviews various aspects related to challenges associated with decision-making at unsignalized intersections with a focus on learning-based schemes. We discuss these schemes in terms of the tackled driving scenario, the involved challenges, the proposed learning-based designs and the validation in simulations and real-world environments. Based on our discussion and investigation, we found that research efforts are still required to tackle the real-world challenges of unsignalized intersection-traversal problem with guarantees of safety and feasibility.
 
 Ultimately, the decision making schemes that were reviewed have been proposed to tackle uncertainties associated with traversing the unsignalized intersection problem. This is commonly modeled as a POMDP due to the unknown intention and future trajectories of intersection users. Environmental uncertainties due to limited sensor range and faulty perception are taken into account while designing occlusion-aware decision making schemes. Furthermore, uncertainties of different driving styles are also considered in developing decision-making schemes for learning optimal crossing policy in multi-agent environments. However, we have observed that specific critical areas have been overlooked, lacking in-depth research. These areas focus on integrating the high-level reasoning principles from the behavioral planning layer with motion planning and control layers, utilizing high-fidelity models to ensure the feasibility of commanded actions. Additionally, there is a need for employing advanced sim-to-real transfer learning techniques to facilitate experimental validation and testing in real-world unsignalized intersection environments. Consequently, we underscored the importance of developing hierarchical decision-making architectures to ensure safety and feasibility. Moreover, we provided suggestions for methods and heuristics that can facilitate real-world driving for testing and validating RL-based models. By incorporating our recommendations, precise and feasible learning-based models can be trained and validated in real-world urban settings.

\section*{Acknowledgment}

This work was supported by NSERC CRD 537104-18, in partnership with General Motors Canada and the SAE AutoDrive Challenge. We would like to thank Dr. Naser Lashgarian Azad and Rowan Dempster for their insightful discussions.

\ifCLASSOPTIONcaptionsoff
  \newpage
\fi



%
\renewcommand*{\bibfont}{\footnotesize}
\bibliographystyle{unsrt}
\bibliography{manuscript}

\begin{thebibliography}{100}

\bibitem{buehler2009darpa}
Martin Buehler, Karl Iagnemma, and Sanjiv Singh.
\newblock {\em The DARPA urban challenge: autonomous vehicles in city traffic}, volume~56.
\newblock springer, 2009.

\bibitem{dikmen2016autonomous}
Murat Dikmen and Catherine~M Burns.
\newblock Autonomous driving in the real world: Experiences with tesla autopilot and summon.
\newblock In {\em Proceedings of the 8th international conference on automotive user interfaces and interactive vehicular applications}, pages 225--228, 2016.

\bibitem{wang2020ethical}
Hong Wang, Amir Khajepour, Dongpu Cao, and Teng Liu.
\newblock Ethical decision making in autonomous vehicles: Challenges and research progress.
\newblock {\em IEEE Intelligent Transportation Systems Magazine}, 2020.

\bibitem{al2020sensorless}
Mohammad Al-Sharman, David Murdoch, Dongpu Cao, Chen Lv, Yahya Zweiri, Derek Rayside, and William Melek.
\newblock A sensorless state estimation for a safety-oriented cyber-physical system in urban driving: deep learning approach.
\newblock {\em IEEE/CAA Journal of Automatica Sinica}, 8(1):169--178, 2020.

\bibitem{kamal2021look}
Md~Abdus~Samad Kamal, Kotaro Hashikura, Tomohisa Hayakawa, Kou Yamada, and Jun-ichi Imura.
\newblock Look-ahead driving schemes for efficient control of automated vehicles on urban roads.
\newblock {\em IEEE Transactions on Vehicular Technology}, 71(2):1280--1292, 2021.

\bibitem{wang2020ethical1}
Hong Wang, Yanjun Huang, Amir Khajepour, Dongpu Cao, and Chen Lv.
\newblock Ethical decision-making platform in autonomous vehicles with lexicographic optimization based model predictive controller.
\newblock {\em IEEE transactions on vehicular technology}, 69(8):8164--8175, 2020.

\bibitem{zhou2023ralacs}
Eddy Zhou, Alex Zhuang, Alikasim Budhwani, Rowan Dempster, Quanquan Li, Mohammad Al-Sharman, Derek Rayside, and William Melek.
\newblock Ralacs: Action recognition in autonomous vehicles using interaction encoding and optical flow, 2023.

\bibitem{kuwata2009real}
Yoshiaki Kuwata, Justin Teo, Gaston Fiore, Sertac Karaman, Emilio Frazzoli, and Jonathan~P How.
\newblock Real-time motion planning with applications to autonomous urban driving.
\newblock {\em IEEE Transactions on control systems technology}, 17(5):1105--1118, 2009.

\bibitem{verma2018vehicle}
Shashwat Verma, You~Hong Eng, Hai~Xun Kong, Hans Andersen, Malika Meghjani, Wei~Kang Leong, Xiaotong Shen, Chen Zhang, Marcelo~H Ang, and Daniela Rus.
\newblock Vehicle detection, tracking and behavior analysis in urban driving environments using road context.
\newblock In {\em 2018 IEEE International Conference on Robotics and Automation (ICRA)}, pages 1413--1420. IEEE, 2018.

\bibitem{senanayake2020directional}
Ransalu Senanayake, Maneekwan Toyungyernsub, Mingyu Wang, Mykel~J Kochenderfer, and Mac Schwager.
\newblock Directional primitives for uncertainty-aware motion estimation in urban environments.
\newblock In {\em 2020 IEEE 23rd International Conference on Intelligent Transportation Systems (ITSC)}, pages 1--6. IEEE, 2020.

\bibitem{yang2023towards}
Kai Yang, Xiaolin Tang, Sen Qiu, Shufeng Jin, Zichun Wei, and Hong Wang.
\newblock Towards robust decision-making for autonomous driving on highway.
\newblock {\em IEEE Transactions on Vehicular Technology}, 2023.

\bibitem{tang2022highway}
Xiaolin Tang, Bing Huang, Teng Liu, and Xianke Lin.
\newblock Highway decision-making and motion planning for autonomous driving via soft actor-critic.
\newblock {\em IEEE Transactions on Vehicular Technology}, 2022.

\bibitem{xu2022integrated}
Can Xu, Wanzhong Zhao, Jinqiang Liu, Chunyan Wang, and Chen Lv.
\newblock An integrated decision-making framework for highway autonomous driving using combined learning and rule-based algorithm.
\newblock {\em IEEE Transactions on Vehicular Technology}, 2022.

\bibitem{li2015real}
Xiaohui Li, Zhenping Sun, Dongpu Cao, Zhen He, and Qi~Zhu.
\newblock Real-time trajectory planning for autonomous urban driving: Framework, algorithms, and verifications.
\newblock {\em IEEE/ASME Transactions on mechatronics}, 21(2):740--753, 2015.

\bibitem{gilroy2019overcoming}
Shane Gilroy, Edward Jones, and Martin Glavin.
\newblock Overcoming occlusion in the automotive environment-a review.
\newblock {\em IEEE Transactions on Intelligent Transportation Systems}, 2019.

\bibitem{huang2019uncertainty}
Xin Huang, Stephen~G McGill, Brian~C Williams, Luke Fletcher, and Guy Rosman.
\newblock Uncertainty-aware driver trajectory prediction at urban intersections.
\newblock In {\em 2019 International Conference on Robotics and Automation (ICRA)}, pages 9718--9724. IEEE, 2019.

\bibitem{dempster2022drg}
Rowan Dempster, Mohammad Al-Sharman, Yeshu Jain, Jeffery Li, Derek Rayside, and William Melek.
\newblock Drg: A dynamic relation graph for unified prior-online environment modeling in urban autonomous driving.
\newblock In {\em 2022 International Conference on Robotics and Automation (ICRA)}, pages 8054--8060. IEEE, 2022.

\bibitem{choi2010crash}
Eun-Ha Choi.
\newblock Crash factors in intersection-related crashes: An on-scene perspective.
\newblock Technical report, 2010.

\bibitem{GM}
Michael Wayland.
\newblock Gm’s cruise begins testing autonomous vehicles without human drivers in san francisco, 2020.

\bibitem{Hu_2023_CVPR}
Yihan Hu, Jiazhi Yang, Li~Chen, Keyu Li, Chonghao Sima, Xizhou Zhu, Siqi Chai, Senyao Du, Tianwei Lin, Wenhai Wang, Lewei Lu, Xiaosong Jia, Qiang Liu, Jifeng Dai, Yu~Qiao, and Hongyang Li.
\newblock Planning-oriented autonomous driving.
\newblock In {\em Proceedings of the IEEE/CVF Conference on Computer Vision and Pattern Recognition (CVPR)}, pages 17853--17862, June 2023.

\bibitem{chitta2022transfuser}
Kashyap Chitta, Aditya Prakash, Bernhard Jaeger, Zehao Yu, Katrin Renz, and Andreas Geiger.
\newblock Transfuser: Imitation with transformer-based sensor fusion for autonomous driving.
\newblock {\em IEEE Transactions on Pattern Analysis and Machine Intelligence}, 2022.

\bibitem{chen2019autonomous}
Yimin Chen, Jingqiang Zha, and Junmin Wang.
\newblock An autonomous t-intersection driving strategy considering oncoming vehicles based on connected vehicle technology.
\newblock {\em IEEE/ASME Transactions on Mechatronics}, 24(6):2779--2790, 2019.

\bibitem{hawke2020urban}
Jeffrey Hawke, Richard Shen, Corina Gurau, Siddharth Sharma, Daniele Reda, Nikolay Nikolov, Przemys{\l}aw Mazur, Sean Micklethwaite, Nicolas Griffiths, Amar Shah, et~al.
\newblock Urban driving with conditional imitation learning.
\newblock In {\em 2020 IEEE International Conference on Robotics and Automation (ICRA)}, pages 251--257. IEEE, 2020.

\bibitem{schwarting2018planning}
Wilko Schwarting, Javier Alonso-Mora, and Daniela Rus.
\newblock Planning and decision-making for autonomous vehicles.
\newblock {\em Annual Review of Control, Robotics, and Autonomous Systems}, 2018.

\bibitem{TaxonomyAD}
Taxonomy and definitions for terms related to driving automation systems for on-road motor vehicles.
\newblock SAE, 2016.

\bibitem{khattak2020exploratory}
Zulqarnain~H Khattak, Michael~D Fontaine, and Brian~L Smith.
\newblock Exploratory investigation of disengagements and crashes in autonomous vehicles under mixed traffic: An endogenous switching regime framework.
\newblock {\em IEEE Transactions on Intelligent Transportation Systems}, 2020.

\bibitem{wang2019crash}
Hong Wang, Yanjun Huang, Amir Khajepour, Yubiao Zhang, Yadollah Rasekhipour, and Dongpu Cao.
\newblock Crash mitigation in motion planning for autonomous vehicles.
\newblock {\em IEEE Transactions on Intelligent Transportation Systems}, 20(9):3313--3323, 2019.

\bibitem{pruekprasert2019game}
Sasinee Pruekprasert, J{\'e}r{\'e}my Dubut, Xiaoyi Zhang, Chao Huang, and Masako Kishida.
\newblock A game theoretic approach to decision making for multiple vehicles at roundabout.
\newblock {\em arXiv preprint arXiv:1904.06224}, 2019.

\bibitem{li2020deep}
Guofa Li, Shenglong Li, Shen Li, Yechen Qin, Dongpu Cao, Xingda Qu, and Bo~Cheng.
\newblock Deep reinforcement learning enabled decision-making for autonomous driving at intersections.
\newblock {\em Automotive Innovation}, 3(4):374--385, 2020.

\bibitem{huang2020diversitygan}
Xin Huang, Stephen~G McGill, Jonathan~A DeCastro, Luke Fletcher, John~J Leonard, Brian~C Williams, and Guy Rosman.
\newblock Diversitygan: Diversity-aware vehicle motion prediction via latent semantic sampling.
\newblock {\em IEEE Robotics and Automation Letters}, 5(4):5089--5096, 2020.

\bibitem{hubmann2018automated}
Constantin Hubmann, Jens Schulz, Marvin Becker, Daniel Althoff, and Christoph Stiller.
\newblock Automated driving in uncertain environments: Planning with interaction and uncertain maneuver prediction.
\newblock {\em IEEE Transactions on Intelligent Vehicles}, 3(1):5--17, 2018.

\bibitem{hafner2013cooperative}
Michael~R Hafner, Drew Cunningham, Lorenzo Caminiti, and Domitilla Del~Vecchio.
\newblock Cooperative collision avoidance at intersections: Algorithms and experiments.
\newblock {\em IEEE Transactions on Intelligent Transportation Systems}, 14(3):1162--1175, 2013.

\bibitem{wu2020cooperative}
Tianhao Wu, Mingzhi Jiang, and Lin Zhang.
\newblock Cooperative multiagent deep deterministic policy gradient (comaddpg) for intelligent connected transportation with unsignalized intersection.
\newblock {\em Mathematical Problems in Engineering}, 2020, 2020.

\bibitem{wang2021digital}
Ziran Wang, Kyungtae Han, and Prashant Tiwari.
\newblock Digital twin-assisted cooperative driving at non-signalized intersections.
\newblock {\em arXiv preprint arXiv:2105.01357}, 2021.

\bibitem{tian2020game}
Ran Tian, Nan Li, Ilya Kolmanovsky, Yildiray Yildiz, and Anouck~R Girard.
\newblock Game-theoretic modeling of traffic in unsignalized intersection network for autonomous vehicle control verification and validation.
\newblock {\em IEEE Transactions on Intelligent Transportation Systems}, 2020.

\bibitem{li2020game}
Nan Li, Yu~Yao, Ilya Kolmanovsky, Ella Atkins, and Anouck~R Girard.
\newblock Game-theoretic modeling of multi-vehicle interactions at uncontrolled intersections.
\newblock {\em IEEE Transactions on Intelligent Transportation Systems}, 2020.

\bibitem{hubmann2017decision}
Constantin Hubmann, Marvin Becker, Daniel Althoff, David Lenz, and Christoph Stiller.
\newblock Decision making for autonomous driving considering interaction and uncertain prediction of surrounding vehicles.
\newblock In {\em 2017 IEEE Intelligent Vehicles Symposium (IV)}, pages 1671--1678. IEEE, 2017.

\bibitem{qiao2018automatically}
Zhiqian Qiao, Katharina Muelling, John~M Dolan, Praveen Palanisamy, and Priyantha Mudalige.
\newblock Automatically generated curriculum based reinforcement learning for autonomous vehicles in urban environment.
\newblock In {\em 2018 IEEE Intelligent Vehicles Symposium (IV)}, pages 1233--1238. IEEE, 2018.

\bibitem{silver2018general}
David Silver, Thomas Hubert, Julian Schrittwieser, Ioannis Antonoglou, Matthew Lai, Arthur Guez, Marc Lanctot, Laurent Sifre, Dharshan Kumaran, Thore Graepel, et~al.
\newblock A general reinforcement learning algorithm that masters chess, shogi, and go through self-play.
\newblock {\em Science}, 362(6419):1140--1144, 2018.

\bibitem{irshayyid2024review}
Ali Irshayyid, Jun Chen, and Guojiang Xiong.
\newblock A review on reinforcement learning-based highway autonomous vehicle control.
\newblock {\em Green Energy and Intelligent Transportation}, page 100156, 2024.

\bibitem{aradi2020survey}
Szil{\'a}rd Aradi.
\newblock Survey of deep reinforcement learning for motion planning of autonomous vehicles.
\newblock {\em IEEE Transactions on Intelligent Transportation Systems}, 23(2):740--759, 2020.

\bibitem{chen2024deep}
Yiyang Chen, Chao Ji, Yunrui Cai, Tong Yan, and Bo~Su.
\newblock Deep reinforcement learning in autonomous car path planning and control: A survey.
\newblock {\em arXiv preprint arXiv:2404.00340}, 2024.

\bibitem{wu2024recent}
Jingda Wu, Chao Huang, Hailong Huang, Chen Lv, Yuntong Wang, and Fei-Yue Wang.
\newblock Recent advances in reinforcement learning-based autonomous driving behavior planning: A survey.
\newblock {\em Transportation Research Part C: Emerging Technologies}, 164:104654, 2024.

\bibitem{zhang2024survey}
Kaiwen Zhang, Zhiyong Cui, and Wanjing Ma.
\newblock A survey on reinforcement learning-based control for signalized intersections with connected automated vehicles.
\newblock {\em Transport Reviews}, pages 1--22, 2024.

\bibitem{muller2022motion}
Johannes M{\"u}ller, Jan Strohbeck, Martin Herrmann, and Michael Buchholz.
\newblock Motion planning for connected automated vehicles at occluded intersections with infrastructure sensors.
\newblock {\em IEEE Transactions on Intelligent Transportation Systems}, 2022.

\bibitem{li2021planning}
Shen Li, Keqi Shu, Chaoyi Chen, and Dongpu Cao.
\newblock Planning and decision-making for connected autonomous vehicles at road intersections: A review.
\newblock {\em Chinese Journal of Mechanical Engineering}, 34(1):1--18, 2021.

\bibitem{golchoubian2023pedestrian}
Mahsa Golchoubian, Moojan Ghafurian, Kerstin Dautenhahn, and Nasser~Lashgarian Azad.
\newblock Pedestrian trajectory prediction in pedestrian-vehicle mixed environments: A systematic review.
\newblock {\em IEEE Transactions on Intelligent Transportation Systems}, 2023.

\bibitem{paden2016survey}
Brian Paden, Michal {\v{C}}{\'a}p, Sze~Zheng Yong, Dmitry Yershov, and Emilio Frazzoli.
\newblock A survey of motion planning and control techniques for self-driving urban vehicles.
\newblock {\em IEEE Transactions on intelligent vehicles}, 1(1):33--55, 2016.

\bibitem{broumi2016applying}
Said Broumi, Assia Bakal, Mohamed Talea, Florentin Smarandache, and Luige Vladareanu.
\newblock Applying dijkstra algorithm for solving neutrosophic shortest path problem.
\newblock In {\em 2016 International conference on advanced mechatronic systems (ICAMechS)}, pages 412--416. IEEE, 2016.

\bibitem{lavalle2006planning}
Steven~M LaValle.
\newblock {\em Planning algorithms}.
\newblock Cambridge university press, 2006.

\bibitem{liu2019intelligent}
Qin Liu, Panlin Hou, Guojun Wang, Tao Peng, and Shaobo Zhang.
\newblock Intelligent route planning on large road networks with efficiency and privacy.
\newblock {\em Journal of Parallel and Distributed Computing}, 133:93--106, 2019.

\bibitem{li2019traffic}
Jinglin Li, Dawei Fu, Quan Yuan, Haohan Zhang, Kaihui Chen, Shu Yang, and Fangchun Yang.
\newblock A traffic prediction enabled double rewarded value iteration network for route planning.
\newblock {\em IEEE Transactions on Vehicular Technology}, 68(5):4170--4181, 2019.

\bibitem{ahmad2019real}
Awais Ahmad, Sadia Din, Anand Paul, Gwanggil Jeon, Moayad Aloqaily, and Mudassar Ahmad.
\newblock Real-time route planning and data dissemination for urban scenarios using the internet of things.
\newblock {\em IEEE Wireless Communications}, 26(6):50--55, 2019.

\bibitem{shirazi2016looking}
Mohammad~Shokrolah Shirazi and Brendan~Tran Morris.
\newblock Looking at intersections: a survey of intersection monitoring, behavior and safety analysis of recent studies.
\newblock {\em IEEE Transactions on Intelligent Transportation Systems}, 18(1):4--24, 2016.

\bibitem{dresner2007sharing}
Kurt~M Dresner and Peter Stone.
\newblock Sharing the road: Autonomous vehicles meet human drivers.
\newblock In {\em Ijcai}, volume~7, pages 1263--1268, 2007.

\bibitem{alonso2011autonomous}
Javier Alonso, Vicente Milan{\'e}s, Joshu{\'e} P{\'e}rez, Enrique Onieva, Carlos Gonz{\'a}lez, and Teresa De~Pedro.
\newblock Autonomous vehicle control systems for safe crossroads.
\newblock {\em Transportation research part C: emerging technologies}, 19(6):1095--1110, 2011.

\bibitem{aoude2010threat}
Georges~S Aoude, Brandon~D Luders, Kenneth~KH Lee, Daniel~S Levine, and Jonathan~P How.
\newblock Threat assessment design for driver assistance system at intersections.
\newblock In {\em 13th International IEEE Conference on Intelligent Transportation Systems}, pages 1855--1862. IEEE, 2010.

\bibitem{okamoto2017driver}
Kazuhide Okamoto, Karl Berntorp, and Stefano Di~Cairano.
\newblock Driver intention-based vehicle threat assessment using random forests and particle filtering.
\newblock {\em IFAC-PapersOnLine}, 50(1):13860--13865, 2017.

\bibitem{li2020threat}
Yang Li, Yang Zheng, Bernhard Morys, Shuyue Pan, Jianqiang Wang, and Keqiang Li.
\newblock Threat assessment techniques in intelligent vehicles: A comparative survey.
\newblock {\em IEEE Intelligent Transportation Systems Magazine}, 2020.

\bibitem{orzechowski2018tackling}
Piotr~F Orzechowski, Annika Meyer, and Martin Lauer.
\newblock Tackling occlusions \& limited sensor range with set-based safety verification.
\newblock In {\em 2018 21st International Conference on Intelligent Transportation Systems (ITSC)}, pages 1729--1736. IEEE, 2018.

\bibitem{jeong2019target}
Yonghwan Jeong and Kyongsu Yi.
\newblock Target vehicle motion prediction-based motion planning framework for autonomous driving in uncontrolled intersections.
\newblock {\em IEEE Transactions on Intelligent Transportation Systems}, 2019.

\bibitem{iranitalab2017comparison}
Amirfarrokh Iranitalab and Aemal Khattak.
\newblock Comparison of four statistical and machine learning methods for crash severity prediction.
\newblock {\em Accident Analysis \& Prevention}, 108:27--36, 2017.

\bibitem{osa2018algorithmic}
Takayuki Osa, Joni Pajarinen, Gerhard Neumann, J~Andrew Bagnell, Pieter Abbeel, and Jan Peters.
\newblock An algorithmic perspective on imitation learning.
\newblock {\em arXiv preprint arXiv:1811.06711}, 2018.

\bibitem{sutton2018reinforcement}
Richard~S Sutton and Andrew~G Barto.
\newblock {\em Reinforcement learning: An introduction}.
\newblock MIT press, 2018.

\bibitem{wang2019deterministic}
Yuanda Wang, Jia Sun, Haibo He, and Changyin Sun.
\newblock Deterministic policy gradient with integral compensator for robust quadrotor control.
\newblock {\em IEEE Transactions on Systems, Man, and Cybernetics: Systems}, 2019.

\bibitem{9144488}
K.~{Zhou}, S.~{Song}, A.~{Xue}, K.~{You}, and H.~{Wu}.
\newblock Smart train operation algorithms based on expert knowledge and reinforcement learning.
\newblock {\em IEEE Transactions on Systems, Man, and Cybernetics: Systems}, pages 1--12, 2020.

\bibitem{de2018integrating}
Tim de~Bruin, Jens Kober, Karl Tuyls, and Robert Babu{\v{s}}ka.
\newblock Integrating state representation learning into deep reinforcement learning.
\newblock {\em IEEE Robotics and Automation Letters}, 3(3):1394--1401, 2018.

\bibitem{chen12019deep}
Jiahong Chen, Tongxin Shu, Teng Li, and Clarence~W de~Silva.
\newblock Deep reinforced learning tree for spatiotemporal monitoring with mobile robotic wireless sensor networks.
\newblock {\em IEEE Transactions on Systems, Man, and Cybernetics: Systems}, 2019.

\bibitem{liu2019integrated}
Jinxin Liu, Yugong Luo, Hui Xiong, Tinghan Wang, Heye Huang, and Zhihua Zhong.
\newblock An integrated approach to probabilistic vehicle trajectory prediction via driver characteristic and intention estimation.
\newblock In {\em 2019 IEEE Intelligent Transportation Systems Conference (ITSC)}, pages 3526--3532. IEEE, 2019.

\bibitem{wang2019trajectory}
Yijing Wang, Zhengxuan Liu, Zhiqiang Zuo, Zheng Li, Li~Wang, and Xiaoyuan Luo.
\newblock Trajectory planning and safety assessment of autonomous vehicles based on motion prediction and model predictive control.
\newblock {\em IEEE Transactions on Vehicular Technology}, 68(9):8546--8556, 2019.

\bibitem{trentin2021interaction}
Vinicius Trentin, Antonio Artu{\~n}edo, Jorge Godoy, and Jorge Villagra.
\newblock Interaction-aware intention estimation at roundabouts.
\newblock {\em IEEE Access}, 9:123088--123102, 2021.

\bibitem{yang2018scene}
Shun Yang, Wenshuo Wang, Chang Liu, and Weiwen Deng.
\newblock Scene understanding in deep learning-based end-to-end controllers for autonomous vehicles.
\newblock {\em IEEE Transactions on Systems, Man, and Cybernetics: Systems}, 49(1):53--63, 2018.

\bibitem{yoo2021virtual}
Jinsoo~Michael Yoo, Yonghwan Jeong, and Kyongsu Yi.
\newblock Virtual target-based longitudinal motion planning of autonomous vehicles at urban intersections: Determining control inputs of acceleration with human driving characteristic-based constraints.
\newblock {\em IEEE Vehicular Technology Magazine}, 16(3):38--46, 2021.

\bibitem{hoel2020reinforcement}
Carl-Johan Hoel, Tommy Tram, and Jonas Sj{\"o}berg.
\newblock Reinforcement learning with uncertainty estimation for tactical decision-making in intersections.
\newblock In {\em 2020 IEEE 23rd international conference on intelligent transportation systems (ITSC)}, pages 1--7. IEEE, 2020.

\bibitem{liu2019driving}
Yonggang Liu, Pan Zhao, Datong Qin, Guang Li, Zheng Chen, and Yi~Zhang.
\newblock Driving intention identification based on long short-term memory and a case study in shifting strategy optimization.
\newblock {\em IEEE Access}, 7:128593--128605, 2019.

\bibitem{trende2021modelling}
Alexander Trende, Anirudh Unni, Jochem Rieger, and Martin Fraenzle.
\newblock Modelling turning intention in unsignalized intersections with bayesian networks.
\newblock In {\em International Conference on Human-Computer Interaction}, pages 289--296. Springer, 2021.

\bibitem{scanlon2016predicting}
John~M Scanlon, Rini Sherony, and Hampton~C Gabler.
\newblock Predicting crash-relevant violations at stop sign--controlled intersections for the development of an intersection driver assistance system.
\newblock {\em Traffic injury prevention}, 17(sup1):59--65, 2016.

\bibitem{doerzaph2007development}
Zachary~Richard Doerzaph.
\newblock {\em Development of a threat assessment algorithm for intersection collision avoidance systems}.
\newblock PhD thesis, Virginia Tech, 2007.

\bibitem{laugier2011probabilistic}
Christian Laugier, Igor~E Paromtchik, Mathias Perrollaz, Mao Yong, John-David Yoder, Christopher Tay, Kamel Mekhnacha, and Amaury N{\`e}gre.
\newblock Probabilistic analysis of dynamic scenes and collision risks assessment to improve driving safety.
\newblock {\em IEEE Intelligent Transportation Systems Magazine}, 3(4):4--19, 2011.

\bibitem{lefevre2012evaluating}
St{\'e}phanie Lef{\`e}vre, Christian Laugier, and Javier Iba{\~n}ez-Guzm{\'a}n.
\newblock Evaluating risk at road intersections by detecting conflicting intentions.
\newblock In {\em 2012 IEEE/RSJ International Conference on Intelligent Robots and Systems}, pages 4841--4846. IEEE, 2012.

\bibitem{zyner2017long}
Alex Zyner, Stewart Worrall, James Ward, and Eduardo Nebot.
\newblock Long short term memory for driver intent prediction.
\newblock In {\em 2017 IEEE Intelligent Vehicles Symposium (IV)}, pages 1484--1489. IEEE, 2017.

\bibitem{phillips2017generalizable}
Derek~J Phillips, Tim~A Wheeler, and Mykel~J Kochenderfer.
\newblock Generalizable intention prediction of human drivers at intersections.
\newblock In {\em 2017 IEEE Intelligent Vehicles Symposium (IV)}, pages 1665--1670. IEEE, 2017.

\bibitem{zyner2018recurrent}
Alex Zyner, Stewart Worrall, and Eduardo Nebot.
\newblock A recurrent neural network solution for predicting driver intention at unsignalized intersections.
\newblock {\em IEEE Robotics and Automation Letters}, 3(3):1759--1764, 2018.

\bibitem{jeong2020surround}
Yonghwan Jeong, Seonwook Kim, and Kyongsu Yi.
\newblock Surround vehicle motion prediction using lstm-rnn for motion planning of autonomous vehicles at multi-lane turn intersections.
\newblock {\em IEEE Open Journal of Intelligent Transportation Systems}, 1:2--14, 2020.

\bibitem{girma2020deep}
Abenezer Girma, Seifemichael Amsalu, Abrham Workineh, Mubbashar Khan, and Abdollah Homaifar.
\newblock Deep learning with attention mechanism for predicting driver intention at intersection.
\newblock {\em arXiv preprint arXiv:2006.05918}, 2020.

\bibitem{pourjafari2023navigating}
Niusha Pourjafari, Amir Ghafari, and Ali Ghaffari.
\newblock Navigating unsignalized intersections: A predictive approach for safe and cautious autonomous driving.
\newblock {\em IEEE Transactions on Intelligent Vehicles}, 2023.

\bibitem{bender2015predicting}
Asher Bender, James~R Ward, Stewart Worrall, and Eduardo~M Nebot.
\newblock Predicting driver intent from models of naturalistic driving.
\newblock In {\em 2015 IEEE 18th International Conference on Intelligent Transportation Systems}, pages 1609--1615. IEEE, 2015.

\bibitem{gadepally2013framework}
Vijay Gadepally, Ashok Krishnamurthy, and Umit Ozguner.
\newblock A framework for estimating driver decisions near intersections.
\newblock {\em IEEE Transactions on Intelligent Transportation Systems}, 15(2):637--646, 2013.

\bibitem{khairdoost2020real}
Nima Khairdoost, Mohsen Shirpour, Michael~A Bauer, and Steven~S Beauchemin.
\newblock Real-time driver maneuver prediction using lstm.
\newblock {\em IEEE Transactions on Intelligent Vehicles}, 5(4):714--724, 2020.

\bibitem{beauchemin2010roadlab}
Steven Beauchemin, M~Bauer, Denis Laurendeau, T~Kowsari, J~Cho, M~Hunter, and O~McCarthy.
\newblock Roadlab: An in-vehicle laboratory for developing cognitive cars.
\newblock In {\em Proc. 23rd Int. Conf. CAINE}, 2010.

\bibitem{azadani2022novel}
Mozhgan~Nasr Azadani and Azzedine Boukerche.
\newblock A novel multimodal vehicle path prediction method based on temporal convolutional networks.
\newblock {\em IEEE Transactions on Intelligent Transportation Systems}, 2022.

\bibitem{zhan2019interaction}
Wei Zhan, Liting Sun, Di~Wang, Haojie Shi, Aubrey Clausse, Maximilian Naumann, Julius Kummerle, Hendrik Konigshof, Christoph Stiller, Arnaud de~La~Fortelle, and Masayoshi Tomizuka.
\newblock Interaction dataset: An international, adversarial and cooperative motion dataset in interactive driving scenarios with semantic maps, 2019.

\bibitem{sarkar2017trajectory}
Atri Sarkar, Krzysztof Czarnecki, Matt Angus, Changjian Li, and Steven Waslander.
\newblock Trajectory prediction of traffic agents at urban intersections through learned interactions.
\newblock In {\em 2017 IEEE 20th International Conference on Intelligent Transportation Systems (ITSC)}, pages 1--8. IEEE, 2017.

\bibitem{brechtel2014probabilistic}
Sebastian Brechtel, Tobias Gindele, and R{\"u}diger Dillmann.
\newblock Probabilistic decision-making under uncertainty for autonomous driving using continuous pomdps.
\newblock In {\em 17th international IEEE conference on intelligent transportation systems (ITSC)}, pages 392--399. IEEE, 2014.

\bibitem{sezer2015towards}
Volkan Sezer, Tirthankar Bandyopadhyay, Daniela Rus, Emilio Frazzoli, and David Hsu.
\newblock Towards autonomous navigation of unsignalized intersections under uncertainty of human driver intent.
\newblock In {\em 2015 IEEE/RSJ International Conference on Intelligent Robots and Systems (IROS)}, pages 3578--3585. IEEE, 2015.

\bibitem{isele2018safe}
David Isele, Alireza Nakhaei, and Kikuo Fujimura.
\newblock Safe reinforcement learning on autonomous vehicles.
\newblock In {\em 2018 IEEE/RSJ International Conference on Intelligent Robots and Systems (IROS)}, pages 1--6. IEEE, 2018.

\bibitem{li2019urban}
Changjian Li and Krzysztof Czarnecki.
\newblock Urban driving with multi-objective deep reinforcement learning.
\newblock In {\em Proceedings of the 18th International Conference on Autonomous Agents and MultiAgent Systems}, pages 359--367. International Foundation for Autonomous Agents and Multiagent Systems, 2019.

\bibitem{bouton2019safe}
Maxime Bouton, Alireza Nakhaei, Kikuo Fujimura, and Mykel~J Kochenderfer.
\newblock Safe reinforcement learning with scene decomposition for navigating complex urban environments.
\newblock In {\em 2019 IEEE Intelligent Vehicles Symposium (IV)}, pages 1469--1476. IEEE, 2019.

\bibitem{bernhard2019addressing}
Julian Bernhard, Stefan Pollok, and Alois Knoll.
\newblock Addressing inherent uncertainty: Risk-sensitive behavior generation for automated driving using distributional reinforcement learning.
\newblock In {\em 2019 IEEE Intelligent Vehicles Symposium (IV)}, pages 2148--2155. IEEE, 2019.

\bibitem{osband2018randomized}
Ian Osband, John Aslanides, and Albin Cassirer.
\newblock Randomized prior functions for deep reinforcement learning.
\newblock In {\em Advances in Neural Information Processing Systems}, pages 8617--8629, 2018.

\bibitem{zhu2017improving}
Pengfei Zhu, Xin Li, Pascal Poupart, and Guanghui Miao.
\newblock On improving deep reinforcement learning for pomdps.
\newblock {\em arXiv preprint arXiv:1704.07978}, 2017.

\bibitem{hong2017deep}
Zhang-Wei Hong, Shih-Yang Su, Tzu-Yun Shann, Yi-Hsiang Chang, and Chun-Yi Lee.
\newblock A deep policy inference q-network for multi-agent systems.
\newblock {\em arXiv preprint arXiv:1712.07893}, 2017.

\bibitem{qiao2018pomdp}
Zhiqian Qiao, Katharina Muelling, John Dolan, Praveen Palanisamy, and Priyantha Mudalige.
\newblock Pomdp and hierarchical options mdp with continuous actions for autonomous driving at intersections.
\newblock In {\em 2018 21st International Conference on Intelligent Transportation Systems (ITSC)}, pages 2377--2382. IEEE, 2018.

\bibitem{igl2018deep}
Maximilian Igl, Luisa Zintgraf, Tuan~Anh Le, Frank Wood, and Shimon Whiteson.
\newblock Deep variational reinforcement learning for pomdps.
\newblock In {\em International Conference on Machine Learning}, pages 2117--2126. PMLR, 2018.

\bibitem{song2016intention}
Weilong Song, Guangming Xiong, and Huiyan Chen.
\newblock Intention-aware autonomous driving decision-making in an uncontrolled intersection.
\newblock {\em Mathematical Problems in Engineering}, 2016, 2016.

\bibitem{shu2021autonomous}
Keqi Shu, Huilong Yu, Xingxin Chen, Shen Li, Long Chen, Qi~Wang, Li~Li, and Dongpu Cao.
\newblock Autonomous driving at intersections: A behavior-oriented critical-turning-point approach for decision making.
\newblock {\em IEEE/ASME Transactions on Mechatronics}, 2021.

\bibitem{barbier2018probabilistic}
Mathieu Barbier, Christian Laugier, Olivier Simonin, and Javier Iba{\~n}ez-Guzm{\'a}n.
\newblock Probabilistic decision-making at road intersections: Formulation and quantitative evaluation.
\newblock In {\em 2018 15th International Conference on Control, Automation, Robotics and Vision (ICARCV)}, pages 795--802. IEEE, 2018.

\bibitem{wang2023learning}
Lingguang Wang, Carlos Fernandez, and Christoph Stiller.
\newblock Learning safe and human-like high-level decisions for unsignalized intersections from naturalistic human driving trajectories.
\newblock {\em IEEE Transactions on Intelligent Transportation Systems}, 2023.

\bibitem{Treiber_2000}
Martin Treiber, Ansgar Hennecke, and Dirk Helbing.
\newblock Congested traffic states in empirical observations and microscopic simulations.
\newblock {\em Physical Review E}, 62(2):1805–1824, August 2000.

\bibitem{yu2019occlusion}
Ming-Yuan Yu, Ram Vasudevan, and Matthew Johnson-Roberson.
\newblock Occlusion-aware risk assessment for autonomous driving in urban environments.
\newblock {\em IEEE Robotics and Automation Letters}, 4(2):2235--2241, 2019.

\bibitem{yu2019risk}
Ming-Yuan Yu, Ram Vasudevan, and Matthew Johnson-Roberson.
\newblock Risk assessment and planning with bidirectional reachability for autonomous driving.
\newblock {\em arXiv preprint arXiv:1909.08059}, 2019.

\bibitem{mcgill2019probabilistic}
Stephen~G McGill, Guy Rosman, Teddy Ort, Alyssa Pierson, Igor Gilitschenski, Brandon Araki, Luke Fletcher, Sertac Karaman, Daniela Rus, and John~J Leonard.
\newblock Probabilistic risk metrics for navigating occluded intersections.
\newblock {\em IEEE Robotics and Automation Letters}, 4(4):4322--4329, 2019.

\bibitem{isele2018navigating}
David Isele, Reza Rahimi, Akansel Cosgun, Kaushik Subramanian, and Kikuo Fujimura.
\newblock Navigating occluded intersections with autonomous vehicles using deep reinforcement learning.
\newblock In {\em 2018 IEEE International Conference on Robotics and Automation (ICRA)}, pages 2034--2039. IEEE, 2018.

\bibitem{lin2019decision}
Xiao Lin, Jiucai Zhang, Jin Shang, Yi~Wang, Hongkai Yu, and Xiaoli Zhang.
\newblock Decision making through occluded intersections for autonomous driving.
\newblock In {\em 2019 IEEE Intelligent Transportation Systems Conference (ITSC)}, pages 2449--2455. IEEE, 2019.

\bibitem{kamran2020risk}
Danial Kamran, Carlos~Fernandez Lopez, Martin Lauer, and Christoph Stiller.
\newblock Risk-aware high-level decisions for automated driving at occluded intersections with reinforcement learning.
\newblock {\em arXiv preprint arXiv:2004.04450}, 2020.

\bibitem{naumann2019safe}
Maximilian Naumann, Hendrik Konigshof, Martin Lauer, and Christoph Stiller.
\newblock Safe but not overcautious motion planning under occlusions and limited sensor range.
\newblock In {\em 2019 IEEE Intelligent Vehicles Symposium (IV)}, pages 140--145. IEEE, 2019.

\bibitem{jeong2019svm}
Yonghwan Jeong, Kyongsu Yi, and Sungmin Park.
\newblock Svm based intention inference and motion planning at uncontrolled intersection.
\newblock {\em IFAC-PapersOnLine}, 52(8):356--361, 2019.

\bibitem{shu2020autonomous}
Keqi Shu, Huilong Yu, Xingxin Chen, Long Chen, Qi~Wang, Li~Li, and Dongpu Cao.
\newblock Autonomous driving at intersections: A critical-turning-point approach for left turns.
\newblock {\em arXiv preprint arXiv:2003.02409}, 2020.

\bibitem{li2021continuous}
Guofa Li, Shenglong Li, Shen Li, and Xingda Qu.
\newblock Continuous decision-making for autonomous driving at intersections using deep deterministic policy gradient.
\newblock {\em IET Intelligent Transport Systems}, 2021.

\bibitem{xiong2016combining}
Xi~Xiong, Jianqiang Wang, Fang Zhang, and Keqiang Li.
\newblock Combining deep reinforcement learning and safety based control for autonomous driving.
\newblock {\em arXiv preprint arXiv:1612.00147}, 2016.

\bibitem{fujimoto2018addressing}
Scott Fujimoto, Herke van Hoof, and David Meger.
\newblock Addressing function approximation error in actor-critic methods, 2018.

\bibitem{9959664}
Shu-Yuan Xu, Xue-Mei Chen, Zi-Jia Wang, Yu-Hui Hu, and Xin-Tong Han.
\newblock Decision-making models for autonomous vehicles at unsignalized intersections based on deep reinforcement learning.
\newblock In {\em 2022 International Conference on Advanced Robotics and Mechatronics (ICARM)}, pages 672--677, 2022.

\bibitem{xu2018learning}
Tianbing Xu, Qiang Liu, Liang Zhao, and Jian Peng.
\newblock Learning to explore with meta-policy gradient, 2018.

\bibitem{mnih2013playing}
Volodymyr Mnih, Koray Kavukcuoglu, David Silver, Alex Graves, Ioannis Antonoglou, Daan Wierstra, and Martin Riedmiller.
\newblock Playing atari with deep reinforcement learning.
\newblock {\em arXiv preprint arXiv:1312.5602}, 2013.

\bibitem{haarnoja2018soft}
Tuomas Haarnoja, Aurick Zhou, Pieter Abbeel, and Sergey Levine.
\newblock Soft actor-critic: Off-policy maximum entropy deep reinforcement learning with a stochastic actor.
\newblock In {\em International conference on machine learning}, pages 1861--1870. PMLR, 2018.

\bibitem{song2021autonomous}
Yunlong Song, HaoChih Lin, Elia Kaufmann, Peter Duerr, and Davide Scaramuzza.
\newblock Autonomous overtaking in gran turismo sport using curriculum reinforcement learning.
\newblock {\em arXiv preprint arXiv:2103.14666}, 2021.

\bibitem{liu2021improved}
Haochen Liu, Zhiyu Huang, and Chen Lv.
\newblock Improved deep reinforcement learning with expert demonstrations for urban autonomous driving.
\newblock {\em arXiv preprint arXiv:2102.09243}, 2021.

\bibitem{yuan2023safe}
Henan Yuan, Penghui Li, Bart van Arem, Liujiang Kang, and Yongqi Dong.
\newblock Safe, efficient, comfort, and energy-saving automated driving through roundabout based on deep reinforcement learning.
\newblock {\em arXiv preprint arXiv:2306.11465}, 2023.

\bibitem{bengio2009curriculum}
Yoshua Bengio, J{\'e}r{\^o}me Louradour, Ronan Collobert, and Jason Weston.
\newblock Curriculum learning.
\newblock In {\em Proceedings of the 26th annual international conference on machine learning}, pages 41--48, 2009.

\bibitem{9981109}
Shivesh Khaitan and John~M. Dolan.
\newblock State dropout-based curriculum reinforcement learning for self-driving at unsignalized intersections.
\newblock In {\em 2022 IEEE/RSJ International Conference on Intelligent Robots and Systems (IROS)}, pages 12219--12224, 2022.

\bibitem{peng2023curriculum}
Zengqi Peng, Xiao Zhou, Yubin Wang, Lei Zheng, Ming Liu, and Jun Ma.
\newblock Curriculum proximal policy optimization with stage-decaying clipping for self-driving at unsignalized intersections, 2023.

\bibitem{9564720}
Hyunki Seong, Chanyoung Jung, Seungwook Lee, and David~Hyunchul Shim.
\newblock Learning to drive at unsignalized intersections using attention-based deep reinforcement learning.
\newblock In {\em 2021 IEEE International Intelligent Transportation Systems Conference (ITSC)}, pages 559--566, 2021.

\bibitem{velickovic2018graph}
Petar Veličković, Guillem Cucurull, Arantxa Casanova, Adriana Romero, Pietro Liò, and Yoshua Bengio.
\newblock Graph attention networks, 2018.

\bibitem{9819830}
Peide Cai, Hengli Wang, Yuxiang Sun, and Ming Liu.
\newblock Dq-gat: Towards safe and efficient autonomous driving with deep q-learning and graph attention networks.
\newblock {\em IEEE Transactions on Intelligent Transportation Systems}, 23(11):21102--21112, 2022.

\bibitem{fortunato2019noisy}
Meire Fortunato, Mohammad~Gheshlaghi Azar, Bilal Piot, Jacob Menick, Ian Osband, Alex Graves, Vlad Mnih, Remi Munos, Demis Hassabis, Olivier Pietquin, Charles Blundell, and Shane Legg.
\newblock Noisy networks for exploration, 2019.

\bibitem{breuer2020opendd}
Antonia Breuer, Jan-Aike Termöhlen, Silviu Homoceanu, and Tim Fingscheidt.
\newblock opendd: A large-scale roundabout drone dataset, 2020.

\bibitem{10160762}
Maximilian Schier, Christoph Reinders, and Bodo Rosenhahn.
\newblock Deep reinforcement learning for autonomous driving using high-level heterogeneous graph representations.
\newblock In {\em 2023 IEEE International Conference on Robotics and Automation (ICRA)}, pages 7147--7153, 2023.

\bibitem{liu2023mtd}
Jiaqi Liu, Peng Hang, Xiao Qi, Jianqiang Wang, and Jian Sun.
\newblock Mtd-gpt: A multi-task decision-making gpt model for autonomous driving at unsignalized intersections.
\newblock In {\em 2023 IEEE 26th International Conference on Intelligent Transportation Systems (ITSC)}, pages 5154--5161. IEEE, 2023.

\bibitem{hester2017deep}
Todd Hester, Matej Vecerik, Olivier Pietquin, Marc Lanctot, Tom Schaul, Bilal Piot, Dan Horgan, John Quan, Andrew Sendonaris, Gabriel Dulac-Arnold, et~al.
\newblock Deep q-learning from demonstrations.
\newblock {\em arXiv preprint arXiv:1704.03732}, 2017.

\bibitem{nair2018overcoming}
Ashvin Nair, Bob McGrew, Marcin Andrychowicz, Wojciech Zaremba, and Pieter Abbeel.
\newblock Overcoming exploration in reinforcement learning with demonstrations.
\newblock In {\em 2018 IEEE International Conference on Robotics and Automation (ICRA)}, pages 6292--6299. IEEE, 2018.

\bibitem{qiao2021behavior}
Zhiqian Qiao, Jeff Schneider, and John~M Dolan.
\newblock Behavior planning at urban intersections through hierarchical reinforcement learning.
\newblock In {\em 2021 IEEE International Conference on Robotics and Automation (ICRA)}, pages 2667--2673. IEEE, 2021.

\bibitem{zare2024survey}
Maryam Zare, Parham~M Kebria, Abbas Khosravi, and Saeid Nahavandi.
\newblock A survey of imitation learning: Algorithms, recent developments, and challenges.
\newblock {\em IEEE Transactions on Cybernetics}, 2024.

\bibitem{cheng2024rethinking}
Jie Cheng, Yingbing Chen, Xiaodong Mei, Bowen Yang, Bo~Li, and Ming Liu.
\newblock Rethinking imitation-based planners for autonomous driving.
\newblock In {\em 2024 IEEE International Conference on Robotics and Automation (ICRA)}, pages 14123--14130. IEEE, 2024.

\bibitem{al2023self}
Mohammad Al-Sharman, Rowan Dempster, Mohamed~A Daoud, Mahmoud Nasr, Derek Rayside, and William Melek.
\newblock Self-learned autonomous driving at unsignalized intersections: A hierarchical reinforced learning approach for feasible decision-making.
\newblock {\em IEEE Transactions on Intelligent Transportation Systems}, 2023.

\bibitem{zhu2021survey}
Zeyu Zhu and Huijing Zhao.
\newblock A survey of deep rl and il for autonomous driving policy learning.
\newblock {\em IEEE Transactions on Intelligent Transportation Systems}, 23(9):14043--14065, 2021.

\bibitem{le2022survey}
Luc Le~Mero, Dewei Yi, Mehrdad Dianati, and Alexandros Mouzakitis.
\newblock A survey on imitation learning techniques for end-to-end autonomous vehicles.
\newblock {\em IEEE Transactions on Intelligent Transportation Systems}, 23(9):14128--14147, 2022.

\bibitem{huang2022efficient}
Zhiyu Huang, Jingda Wu, and Chen Lv.
\newblock Efficient deep reinforcement learning with imitative expert priors for autonomous driving.
\newblock {\em IEEE Transactions on Neural Networks and Learning Systems}, 2022.

\bibitem{krajzewicz2012recent}
Daniel Krajzewicz, Jakob Erdmann, Michael Behrisch, and Laura Bieker.
\newblock Recent development and applications of sumo-simulation of urban mobility.
\newblock {\em International journal on advances in systems and measurements}, 5(3\&4), 2012.

\bibitem{dosovitskiy2017carla}
Alexey Dosovitskiy, German Ros, Felipe Codevilla, Antonio Lopez, and Vladlen Koltun.
\newblock Carla: An open urban driving simulator.
\newblock {\em arXiv preprint arXiv:1711.03938}, 2017.

\bibitem{liu2020decision2}
Teng Liu, Xingyu Mu, Bing Huang, Xiaolin Tang, Fuqing Zhao, Xiao Wang, and Dongpu Cao.
\newblock Decision-making at unsignalized intersection for autonomous vehicles: Left-turn maneuver with deep reinforcement learning.
\newblock {\em arXiv preprint arXiv:2008.06595}, 2020.

\bibitem{bouton2019reinforcement}
Maxime Bouton, Jesper Karlsson, Alireza Nakhaei, Kikuo Fujimura, Mykel~J Kochenderfer, and Jana Tumova.
\newblock Reinforcement learning with probabilistic guarantees for autonomous driving.
\newblock {\em arXiv preprint arXiv:1904.07189}, 2019.

\bibitem{shu2022driving}
Hong Shu, Teng Liu, Xingyu Mu, and Dongpu Cao.
\newblock Driving tasks transfer using deep reinforcement learning for decision-making of autonomous vehicles in unsignalized intersection.
\newblock {\em IEEE Transactions on Vehicular Technology}, 2022.

\bibitem{9922164}
Chi Zhang, Kais Kacem, Gereon Hinz, and Alois Knoll.
\newblock Safe and rule-aware deep reinforcement learning for autonomous driving at intersections.
\newblock In {\em 2022 IEEE 25th International Conference on Intelligent Transportation Systems (ITSC)}, pages 2708--2715, 2022.

\bibitem{liu2022improved}
Haochen Liu, Zhiyu Huang, Jingda Wu, and Chen Lv.
\newblock Improved deep reinforcement learning with expert demonstrations for urban autonomous driving.
\newblock In {\em 2022 IEEE intelligent vehicles symposium (IV)}, pages 921--928. IEEE, 2022.

\bibitem{cai2023rule}
Yingfeng Cai, Rong Zhou, Hai Wang, Xiaoqiang Sun, Long Chen, Yicheng Li, Qingchao Liu, and Youguo He.
\newblock Rule-constrained reinforcement learning control for autonomous vehicle left turn at unsignalized intersection.
\newblock {\em IET Intelligent Transport Systems}, 17(11):2143--2153, 2023.

\bibitem{daoud2022simultaneous}
Mohamed~A Daoud, Mohamed~W Mehrez, Derek Rayside, and William~W Melek.
\newblock Simultaneous feasible local planning and path-following control for autonomous driving.
\newblock {\em IEEE Transactions on Intelligent Transportation Systems}, 2022.

\bibitem{dempster2023real}
Rowan Dempster, Mohammad Al-Sharman, Derek Rayside, and William Melek.
\newblock Real-time unified trajectory planning and optimal control for urban autonomous driving under static and dynamic obstacle constraints.
\newblock In {\em 2023 IEEE International Conference on Robotics and Automation (ICRA)}, pages 10139--10145. IEEE, 2023.

\bibitem{hu2021event}
Chaofang Hu, Lingxue Zhao, and Ge~Qu.
\newblock Event-triggered model predictive adaptive dynamic programming for road intersection path planning of unmanned ground vehicle.
\newblock {\em IEEE Transactions on Vehicular Technology}, 70(11):11228--11243, 2021.

\bibitem{bautista2022autonomous}
Rolando Bautista-Montesano, Renato Galluzzi, Kangrui Ruan, Yongjie Fu, and Xuan Di.
\newblock Autonomous navigation at unsignalized intersections: A coupled reinforcement learning and model predictive control approach.
\newblock {\em Transportation research part C: emerging technologies}, 139:103662, 2022.

\bibitem{wang2021distributed}
Kaizheng Wang, Yafei Wang, Lin Wang, Haiping Du, and Kanghyun Nam.
\newblock Distributed intersection conflict resolution for multiple vehicles considering longitudinal-lateral dynamics.
\newblock {\em IEEE Transactions on Vehicular Technology}, 70(5):4166--4177, 2021.

\bibitem{hamouda2021multi}
Ahmed~H Hamouda, Dalia~M Mahfouz, Catherine~M Elias, and Omar~M Shehata.
\newblock Multi-layer control architecture for unsignalized intersection management via nonlinear mpc and deep reinforcement learning.
\newblock In {\em 2021 IEEE International Intelligent Transportation Systems Conference (ITSC)}, pages 1990--1996. IEEE, 2021.

\bibitem{heidecker2021application}
Florian Heidecker, Jasmin Breitenstein, Kevin R{\"o}sch, Jonas L{\"o}hdefink, Maarten Bieshaar, Christoph Stiller, Tim Fingscheidt, and Bernhard Sick.
\newblock An application-driven conceptualization of corner cases for perception in highly automated driving.
\newblock In {\em 2021 IEEE Intelligent Vehicles Symposium (IV)}, pages 644--651. IEEE, 2021.

\bibitem{pitropov2021canadian}
Matthew Pitropov, Danson~Evan Garcia, Jason Rebello, Michael Smart, Carlos Wang, Krzysztof Czarnecki, and Steven Waslander.
\newblock Canadian adverse driving conditions dataset.
\newblock {\em The International Journal of Robotics Research}, 40(4-5):681--690, 2021.

\bibitem{zhang2023perception}
Yuxiao Zhang, Alexander Carballo, Hanting Yang, and Kazuya Takeda.
\newblock Perception and sensing for autonomous vehicles under adverse weather conditions: A survey.
\newblock {\em ISPRS Journal of Photogrammetry and Remote Sensing}, 196:146--177, 2023.

\bibitem{kiran2020deep}
B~Ravi Kiran, Ibrahim Sobh, Victor Talpaert, Patrick Mannion, Ahmad A~Al Sallab, Senthil Yogamani, and Patrick P{\'e}rez.
\newblock Deep reinforcement learning for autonomous driving: A survey.
\newblock {\em arXiv preprint arXiv:2002.00444}, 2020.

\bibitem{10242366}
Xuemin Hu, Shen Li, Tingyu Huang, Bo~Tang, Rouxing Huai, and Long Chen.
\newblock How simulation helps autonomous driving: A survey of sim2real, digital twins, and parallel intelligence.
\newblock {\em IEEE Transactions on Intelligent Vehicles}, pages 1--20, 2023.

\bibitem{9606868}
Erica Salvato, Gianfranco Fenu, Eric Medvet, and Felice~Andrea Pellegrino.
\newblock Crossing the reality gap: A survey on sim-to-real transferability of robot controllers in reinforcement learning.
\newblock {\em IEEE Access}, 9:153171--153187, 2021.

\bibitem{osinski2020simulation}
B{\l}a{\.z}ej Osi{\'n}ski, Adam Jakubowski, Pawe{\l} Ziecina, Piotr Mi{\l}o{\'s}, Christopher Galias, Silviu Homoceanu, and Henryk Michalewski.
\newblock Simulation-based reinforcement learning for real-world autonomous driving.
\newblock In {\em 2020 IEEE international conference on robotics and automation (ICRA)}, pages 6411--6418. IEEE, 2020.

\bibitem{pan2017virtual}
Xinlei Pan, Yurong You, Ziyan Wang, and Cewu Lu.
\newblock Virtual to real reinforcement learning for autonomous driving.
\newblock {\em arXiv preprint arXiv:1704.03952}, 2017.

\bibitem{peng2018sim}
Xue~Bin Peng, Marcin Andrychowicz, Wojciech Zaremba, and Pieter Abbeel.
\newblock Sim-to-real transfer of robotic control with dynamics randomization.
\newblock In {\em 2018 IEEE international conference on robotics and automation (ICRA)}, pages 1--8. IEEE, 2018.

\bibitem{8202133}
Josh Tobin, Rachel Fong, Alex Ray, Jonas Schneider, Wojciech Zaremba, and Pieter Abbeel.
\newblock Domain randomization for transferring deep neural networks from simulation to the real world.
\newblock In {\em 2017 IEEE/RSJ International Conference on Intelligent Robots and Systems (IROS)}, pages 23--30, 2017.

\bibitem{BHATTACHARYYA201745}
S.P. Bhattacharyya.
\newblock Robust control under parametric uncertainty: An overview and recent results.
\newblock {\em Annual Reviews in Control}, 44:45--77, 2017.

\bibitem{9294396}
Georgios~D. Kontes, Daniel~D. Scherer, Tim Nisslbeck, Janina Fischer, and Christopher Mutschler.
\newblock High-speed collision avoidance using deep reinforcement learning and domain randomization for autonomous vehicles.
\newblock In {\em 2020 IEEE 23rd International Conference on Intelligent Transportation Systems (ITSC)}, pages 1--8, 2020.

\bibitem{Pouyanfar_2019_CVPR_Workshops}
Samira Pouyanfar, Muneeb Saleem, Nikhil George, and Shu-Ching Chen.
\newblock Roads: Randomization for obstacle avoidance and driving in simulation.
\newblock In {\em Proceedings of the IEEE/CVF Conference on Computer Vision and Pattern Recognition (CVPR) Workshops}, June 2019.

\bibitem{amini2020learning}
Alexander Amini, Igor Gilitschenski, Jacob Phillips, Julia Moseyko, Rohan Banerjee, Sertac Karaman, and Daniela Rus.
\newblock Learning robust control policies for end-to-end autonomous driving from data-driven simulation.
\newblock {\em IEEE Robotics and Automation Letters}, 5(2):1143--1150, 2020.

\bibitem{pmlr-v70-pinto17a}
Lerrel Pinto, James Davidson, Rahul Sukthankar, and Abhinav Gupta.
\newblock Robust adversarial reinforcement learning.
\newblock In Doina Precup and Yee~Whye Teh, editors, {\em Proceedings of the 34th International Conference on Machine Learning}, volume~70 of {\em Proceedings of Machine Learning Research}, pages 2817--2826. PMLR, 06--11 Aug 2017.

\bibitem{9994638}
Xiangkun He, Baichuan Lou, Haohan Yang, and Chen Lv.
\newblock Robust decision making for autonomous vehicles at highway on-ramps: A constrained adversarial reinforcement learning approach.
\newblock {\em IEEE Transactions on Intelligent Transportation Systems}, 24(4):4103--4113, 2023.

\bibitem{8794293}
Xinlei Pan, Daniel Seita, Yang Gao, and John Canny.
\newblock Risk averse robust adversarial reinforcement learning.
\newblock In {\em 2019 International Conference on Robotics and Automation (ICRA)}, pages 8522--8528, 2019.

\bibitem{bousmalis2017unsupervised}
Konstantinos Bousmalis, Nathan Silberman, David Dohan, Dumitru Erhan, and Dilip Krishnan.
\newblock Unsupervised pixel-level domain adaptation with generative adversarial networks.
\newblock In {\em Proceedings of the IEEE conference on computer vision and pattern recognition}, pages 3722--3731, 2017.

\bibitem{hu2022sim}
Chuqing Hu, Sinclair Hudson, Martin Ethier, Mohammad Al-Sharman, Derek Rayside, and William Melek.
\newblock Sim-to-real domain adaptation for lane detection and classification in autonomous driving.
\newblock {\em arXiv preprint arXiv:2202.07133}, 2022.

\bibitem{ganin2016domain}
Yaroslav Ganin, Evgeniya Ustinova, Hana Ajakan, Pascal Germain, Hugo Larochelle, Fran{\c{c}}ois Laviolette, Mario Marchand, and Victor Lempitsky.
\newblock Domain-adversarial training of neural networks.
\newblock {\em The Journal of Machine Learning Research}, 17(1):2096--2030, 2016.

\bibitem{bewley2019learning}
Alex Bewley, Jessica Rigley, Yuxuan Liu, Jeffrey Hawke, Richard Shen, Vinh-Dieu Lam, and Alex Kendall.
\newblock Learning to drive from simulation without real world labels.
\newblock In {\em 2019 International Conference on Robotics and Automation (ICRA)}, pages 4818--4824. IEEE, 2019.

\end{thebibliography}

\end{document}